\documentclass[journal]{IEEEtran}
\usepackage{graphicx}
\usepackage{booktabs}
\usepackage{xcolor}
\usepackage{multirow}
\usepackage{subfigure}
\def\mygreen#1{{\color{green}#1}}
\ifCLASSINFOpdf
\else
\fi
\begin{document}
%
\title{Depthwise Non-local Module for Fast Salient Object Detection Using a Single Thread}
%
%
%

\author{Haofeng~Li,
        Guanbin~Li,
        Binbin~Yang,
        Guanqi~Chen,
        Liang~Lin,
        Yizhou~Yu
\thanks{This work was supported in part by the National Natural Science Foundation of China under Grant No.61976250, No.61702565 and No.U1811463, in part by the NSFC-Shenzhen Robotics Projects~(U1613211), in part by the Fundamental Research Funds for the Central Universities under Grant No.18lgpy63, and in part by the National High Level Talents Special Support Plan~(Ten Thousand Talents Program). (Corresponding authors: Guanbin Li and Yizhou Yu).}

\thanks{H. Li and Y. Yu are with the Department
of Computer Science, The University of Hong Kong, Hong Kong. 
(e-mail: lhaof@foxmail.com; yizhouy@acm.org).}
\thanks{G. Li, B. Yang, G. Chen and L. Lin are with the school of Data and Computer Science, Sun Yat-sen University, Guangzhou 510006, China (e-mail: liguanbin@mail.sysu.edu.cn; yangbb3@mail2.sysu.edu.cn; chengq26@mail2.sysu.edu.cn; linliang@ieee.org).}}

%
%

\markboth{Journal of \LaTeX\ Class Files,~Vol.~14, No.~8, August~2015}%
{Shell \MakeLowercase{\textit{et al.}}: Bare Demo of IEEEtran.cls for IEEE Journals}
%



\maketitle

\begin{abstract}
    Recently deep convolutional neural networks have achieved significant success in salient object detection. However, existing state-of-the-art methods require high-end GPUs to achieve real-time performance, which makes them hard to adapt to low-cost or portable devices. Although generic network architectures have been proposed to speed up inference on mobile devices, they are tailored to the task of image classification or semantic segmentation, and struggle to capture intra-channel and inter-channel correlations that are essential for contrast modeling in salient object detection. Motivated by the above observations, we design a new deep learning algorithm for fast salient object detection. The proposed algorithm for the first time achieves competitive accuracy and high inference efficiency simultaneously with a single CPU thread. Specifically, we propose a novel depthwise non-local moudule~(DNL), which implicitly models contrast via harvesting intra-channel and inter-channel correlations in a self-attention manner. In addition, we introduce a depthwise non-local network architecture that incorporates both depthwise non-local modules and inverted residual blocks. Experimental results show that our proposed network attains very competitive accuracy on a wide range of salient object detection datasets while achieving state-of-the-art efficiency among all existing deep learning based algorithms.
\end{abstract}


%
\IEEEpeerreviewmaketitle

\section{Introduction}

%
%
%
%
\IEEEPARstart{S}{alient} object detection, which aims to identify the most visually distinctive objects within an image, has been well studied. Developing an accurate salient object detection model benefits a series of applications, such as person re-identification~\cite{zhao2013unsupervised}, robotic control~\cite{shon2005probabilistic}, object detection~\cite{navalpakkam2006integrated},  visual tracking~\cite{wu2014weighted} and content-aware image editing~\cite{avidan2007seam}. Salient object detection usually serves as a pre-processing component, which not only requires acceptable accuracy but also fast speed and small memory consumption on low-cost devices. Recent deep convolutional neural networks (CNNs) exhibit remarkable performance on many computer vision tasks including salient object detection, due to its strong fitting capacity. In particular, dense labeling methods, which make use of fully convolutional network (FCN) architecture, enjoy high accuracy and efficiency offered by end-to-end training and inference. However, the acceleration of convolution operations is highly dependent on high-performance GPUs, which are typically not supported by mobile devices, embedded devices and low-cost personal computers. Developing a deep learning based salient object detection algorithm, that achieves both fast inference and high-quality results using a single thread, remains a challenging task.

Generic low-cost deep network architectures have been proposed recently for mobile devices. Most of them replace conventional convolutional operators with a combination of depthwise separable convolutions and $1\times1$ convolutions. The inverted residual block~\cite{sandler2018mobilenetv2} is one of such neural network modules based on depthwise separable convolutions. For example, an inverted residual block first expands the feature at each spatial position to a higher dimension, and then independently applies a convolution operation on each channel slice. Such methods demonstrate desired inference speed on CPUs but their prediction quality is far from satisfactory. The most essential reason behind is that these lightweight methods directly discard correlation modeling at different channels and spatial positions. 
Such correlation can be taken as context information, which plays an important role in modeling coherence, contrast and uniqueness for salient object detection. Simply borrowing existing generic lightweight network architectures to salient object detection does obtain high efficiency, but their prediction accuracies are far from competitive.

\begin{figure}[t]
\includegraphics[width=1.0\linewidth]{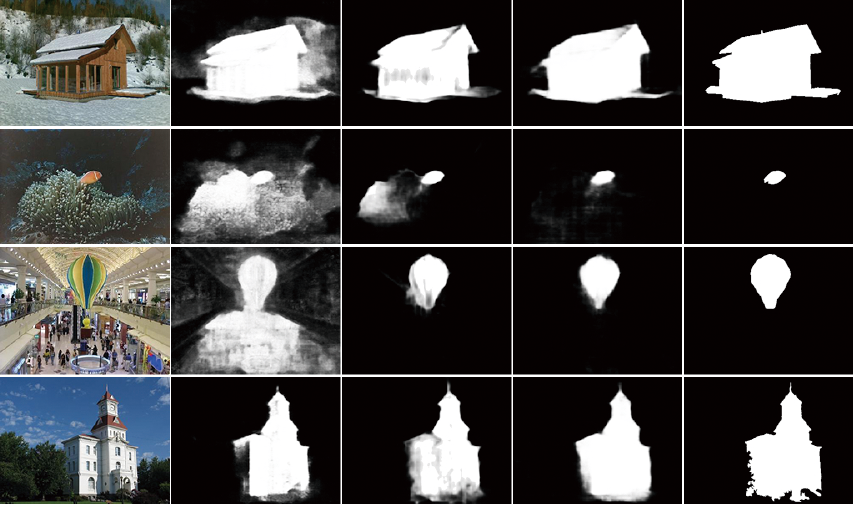}
\leftline{\small \hspace{0.1cm}input image \hspace{0.6cm}UCF \hspace{0.8cm}PAGRN \hspace{0.9cm}ours \hspace{0.9cm}ground}
\leftline{\small \hspace{7.7cm}truth}
   \caption{Comparison among UCF~\cite{zhang2017learning}, PAGRN~\cite{zhang2018progressive} and our DNL algorithm. The proposed algorithm not only achieves practical inference time consumption using a single thread, but also presents saliency maps of competitive quality.}
\label{fig:simpleComp}
\end{figure}

Driven by the above insights, this paper proposes a novel depthwise non-local module, and a fast salient object detection network framework based on the proposed module. This module aims at exploiting relationships among features located at different channel slices or positions. In contrast to traditional convolutional layers that take the vector across all channels at the same spatial position as a feature, the depthwise non-local module considers one column vector or row vector in the same channel as a feature unit, which is described as `depthwise'. The proposed module then learns an attention map via calculating pairwise similarity among non-local features within each sub-region. From the resulting attention map, a feature residual is computed in a self-attention way to update the feature map.

Our proposed module has the following strengths. First, the depthwise non-local module overcomes the limitations of inverted residual blocks by explicitly inferring correlations among features that are neither in the same channel nor in the same spatial position. Second, the DNL module can implicitly model connectivity, contrast and coherence for salient object detection. For example, if image parts are visually similar, their visual features have high attention scores. Thus, their salient features likely update each other and these image parts can be labeled with close saliency values. When a target feature is widely apart from other surrounding features in their latent space, the image region corresponding to the feature may have a different saliency value from other surrounding regions. Third, our proposed module segments an input feature map into sub-regions and only applies self-attention within each sub-region, which lowers its computational cost. As a result, the proposed module adds a very small amount of computation while it considerably enhances the accuracy of a baseline.

This paper has the following contributions.
\begin{itemize}
\item We propose a novel depthwise non-local module, which aims at mining intra-channel and inter-channel correlations in context. The proposed module enhances the fitting capacity of inverted residual blocks at the cost of negligible extra inference time.
\item {We present a fast depthwise non-local neural network, which not only demonstrates state-of-the-art inference speed with a single CPU thread  but also attains competitive detection accuracy (as shown in Fig.~\ref{fig:simpleComp}) among deep learning methods.}
\item We have conducted extensive experiments, which verify the effectiveness and efficiency of the depthwise non-local module and the proposed network framework.
\end{itemize}

\section{Related Work}

\subsection{Salient Object Detection}
Salient object detection can be solved by computing saliency map with prior knowledge and handcrafted features~\cite{cheng2011global,perazzi2012saliency,jiang2013salient,zhang2015minimum,tu2016real,huang2017300,peng2017ahybrid,zhang2017ranking}, or training a deep learning model for prediction~\cite{li2015visual,wang2015deep,li2018contrast,li2016deepsaliency,liu2016dhsnet,luo2017non,wang2018salient,sun2018sgfcn,li2019robust,qin2018hierarchical,chen2018reverse,Li_2018_CVPR,Li_2019_ICCV}. MBS~\cite{zhang2015minimum} exploits the cue that background regions are usually connected to the image boundaries, by computing an approximate minimum barrier distance transform. MST~\cite{tu2016real} employs a similar prior with MBS, but computes an exact distance transform with a minimum spanning tree to measure boundary connectivity. MDC~\cite{huang2017300} suggests that background pixels display low contrast in at least one dimension and proposes minimum directional contrast as raw saliency for each pixel. Priors based methods enjoy real-time efficiency but cannot attain state-of-the-art results. Deep learning based salient object detection models can be roughly divided into two groups, including sparse and dense labeling.  
MDF~\cite{li2015visual} employs deep CNNs to extract multi-scale features and predict saliency values for image segments of different levels. Zeng \textit{et al.}~\cite{zeng2018game} formulate saliency detection as a non-cooperative game, where image regions as players choose to be foreground or background. Zhang \textit{et al.}~\cite{zhang2018absorbing} convert an image into a sparsely-connected graph of regions, and compute saliency via an absorbing Markov chain. Qin \textit{et al.}~\cite{qin2018hierarchical} develop Single-layer Cellular Automata (SCA) that can utilize the intrinsic correlations of similar image patches to locate salient objects, based on deep learning features.
These sparse labeling methods require dividing an input image into hundreds of segments and estimating saliency value for each segment, which is not efficient for real-time applications. To name a few dense labeling methods, DSS~\cite{HouPami18Dss} introduces a series of side output layers and short connections to combine the advantages of low-level and high-level features. 
PAGRN~\cite{zhang2018progressive} is a progressive attention driven framework based on multi-path recurrent feedback. Wang \textit{et al.}~\cite{wang2018detect} propose a global recurrent localization network to locate salient objects, and a local boundary refinement network to capture pixel relations. Liu \textit{et al.}~\cite{liu19PoolNet} integrate a global guidance module and a feature aggregation module into a U-shape architecture.

\subsection{Fast Convolutional Neural Network}
Designing efficient and lightweight neural networks~\cite{ngiam2010tiled,han2015learning,liu2015sparse,sindhwani2015structured,rastegari2016xnor,zhang2016accelerating,wang2017factorized,he2017channel,Zhang_2018_CVPR} has recently become popular in the community. Han \textit{et al.}~\cite{han2015learning} propose a network pruning pipeline that is first trained to learn which connections are important, and then discards the unimportant connections. Factorized Convolutional Neural Networks~\cite{wang2017factorized} unravel the 3D convolution operation in a convolution layer as spatial convolutions in each channel and a linear projection across channels, to reduce the computation.
He \textit{et al.}~\cite{he2017channel} introduce a channel pruning method which alternatively select the most representative channels based on a LASSO regression, and reconstruct the output feature maps with linear least squares.
ShuffleNet~\cite{Zhang_2018_CVPR} proposes a pointwise group convolution that separates convolution filters into groups, and `channel shuffle' that permutes the channels in a group. MobileNetV2~\cite{sandler2018mobilenetv2} utilizes an inverted residual structure that is composed of two pointwise convolutions and a depthwise separable convolution. Some existing state-of-the-art fast deep neural networks employ depthwise separable convolutions that are lightweight but lack intra-channel and inter-channel correlations mining.

\subsection{Self-Attention and Non-local Modeling}
Computational models that exploit pairwise similarities as self-attention scores among dense positions or nodes within some non-local regions, have been widely studied in the field of natural language processing and computer vision~\cite{battaglia2016interaction,buades2005non,He_2019_AAAI,efros1999texture,krahenbuhl2011efficient,santoro2017simple,vaswani2017attention,Wang_2018_CVPR,watters2017visual}. Self-attention model for machine translation~\cite{vaswani2017attention} learns a feature for some node by attending to all other nodes in the same sequence and taking weighted summation of their embedded features in some latent space. Non-local means algorithm~\cite{buades2005non} denoises an image by replacing a pixel with a non-local averaging of all pixels in the image. The non-local averaging utilizes similarity between pixels as weights. Block-matching 3D (BM3D)~\cite{dabov2007image} applies collaborative filtering on a group of similar non-local image patches and achieves competitive image denoising results, even compared to deep learning based methods. Efros and Leung~\cite{efros1999texture} synthesize texture by growing one pixel at a time. They determine pixel value by locating all image patches matching with the target position to fill. Dense conditional random field (CRF)~\cite{krahenbuhl2011efficient} models long-range dependencies by introducing a pairwise energy term that is weighted by the similarity of two nodes. Li et al.~\cite{li2018non} propose a non-locally enhanced encoder-decoder network which can learn more accurate feature for rain steaks and preserve better image details during de-raining. Besides from low-level tasks, Wang \emph{et al.}~\cite{Wang_2018_CVPR} propose a non-local neural network that can harvest long-range spatiotemporal relations for high-level problems, such as video classifications. Such models can learn features from long-range dependencies that are potential to model contrast and coherency in salient object detection. Most existing non-local models work in the spatial dimensions of images or the spatiotemporal dimensions of videos.
\begin{figure}[t]
\centering
\includegraphics[width=1.0\linewidth]{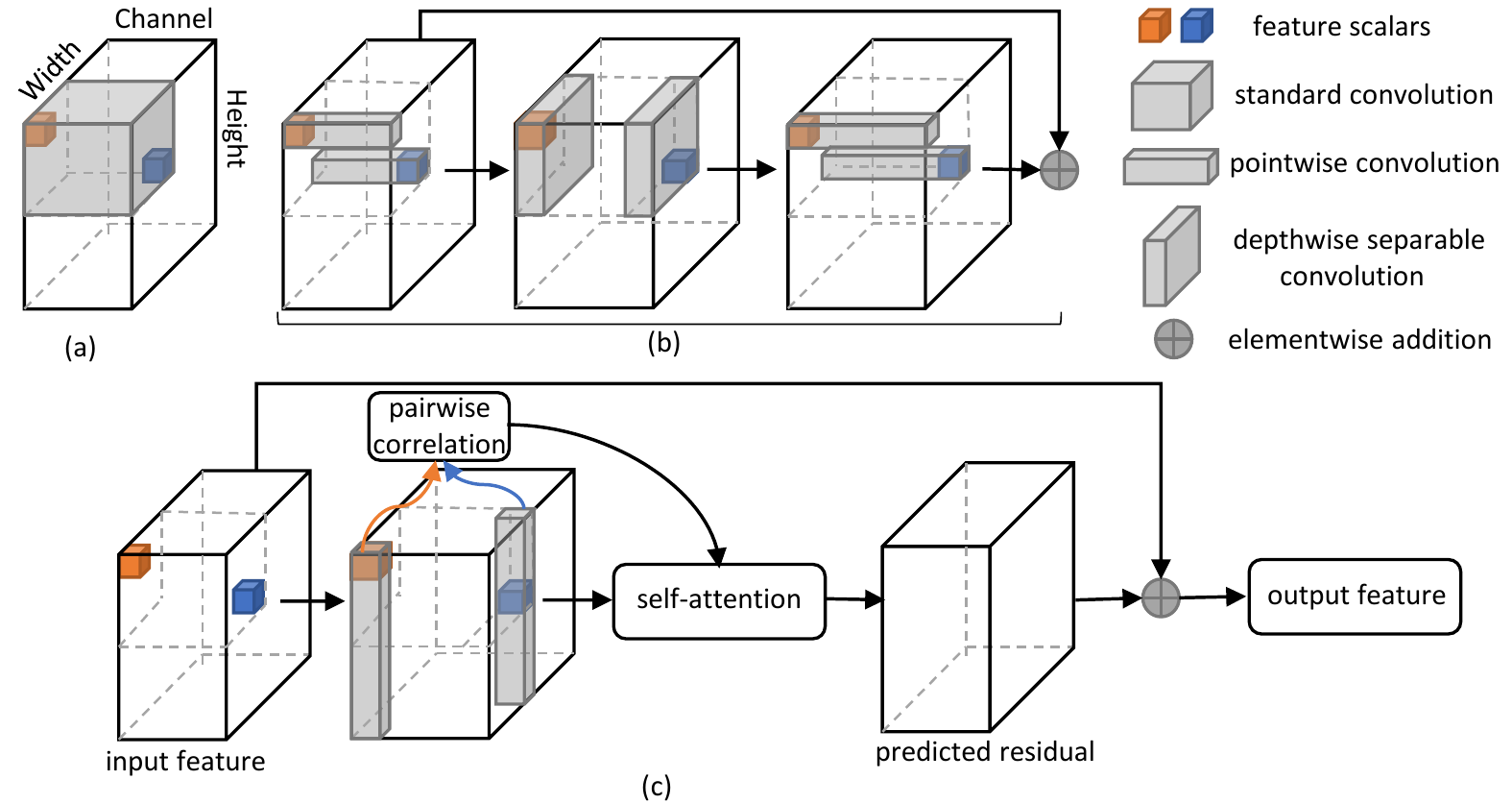}
\caption{(a) Standard Convolution. (b) Inverted Residual. Before the element-wise addition in inverted residual, the shapes of the two feature maps are aligned with each other. Notice that the correlation of two spatially close scalars (in orange and blue colors respectively) is not directly captured in an inverted residual, while a standard convolution can predict outputs based on them together. (c) Our proposed depthwise non-local module. The proposed module can capture the correlation between two feature vectors, which are located at different channels and spatial positions.}
\label{fig:DNLmotivation}
\end{figure}
\begin{figure*}[!ht]
\centering
\includegraphics[width=1.0\linewidth]{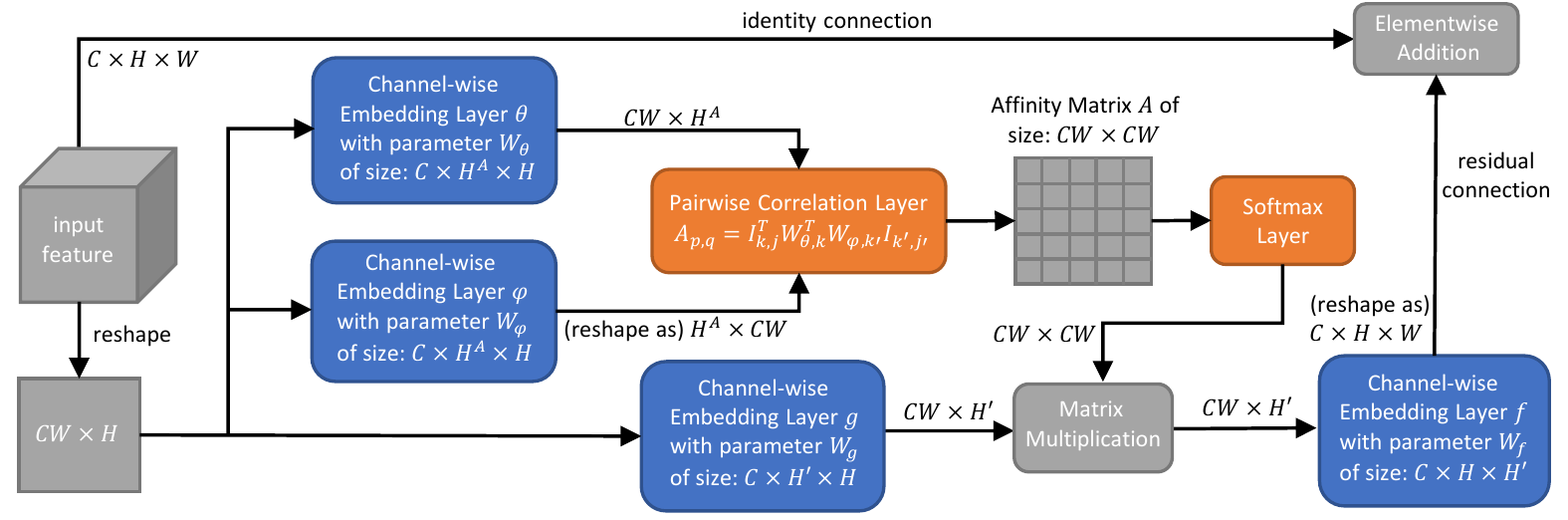}
\caption{Vanilla Depthwise Non-Local Module (vertical-split). The proposed module contains four channel-wise embedding layers, a pairwise correlation layer and a softmax layer. The softmax layer normalizes each row of the affinity matrix $A$. The matrix multiplication serves as taking weighted summation of the output of Layer $g$, with the softmax output as attention weights.}
\label{fig:DNLmodule}
\end{figure*}
\section{Method}

\subsection{Depthwise Non-local Module}
In this section, we introduce a novel depthwise non-local module, that efficiently enhances inverted residual blocks with channel-wise coherence and contrast learning. Inverted residual is an efficient neural network block built on top of depthwise separable convolutions. In the following, we briefly review inverted residual blocks and depthwise separable convolutions as preliminaries. Let $I$ be a $C \times H\times W$ input feature map. $C$, $H$ and $W$ denotes the channel number (depth), the height and the width of $I$ respectively. $k$, $i$ and $j$ are used as the indices of depth, height and width respectively. For example, $I_{i,j}$ is a vector of length $C$ and $I_{k,i}$ has $W$ elements. Consider a regular $K \times K$ convolution layer with $C$ output channels. Its total number of weights is $CK^2 \times C$. The time complexity of applying convolution at one position is $C^2K^2$. As for a depthwise separable convolution layer with $C$ output channels, it has $C$ independent convolution kernels, each of which has a total of $K^2 \times 1$ weights. Performing depthwise separable convolution at one position costs $CK^2$. As shown in Fig.~\ref{fig:DNLmotivation}(b), an inverted residual block is composed of a $1\times 1$ convolution, a depthwise convolution and another $1\times 1$ convolution. These two $1 \times 1$ convolutions aim at aggregating features across channels. However, for a pair of positions (the orange and blue feature scalars in Fig.~\ref{fig:DNLmotivation}), $I_{k,i,j}$ and $I_{k',i',j'}$ ($k \neq k'$ and $(i,j) \neq (i',j')$), their correlation cannot be directly captured by the depthwise separable convolution or the pointwise $1 \times 1$ convolutions even when they are located within each other's neighborhood.

\begin{figure*}[!ht]
\centering
\includegraphics[width=1.0\linewidth]{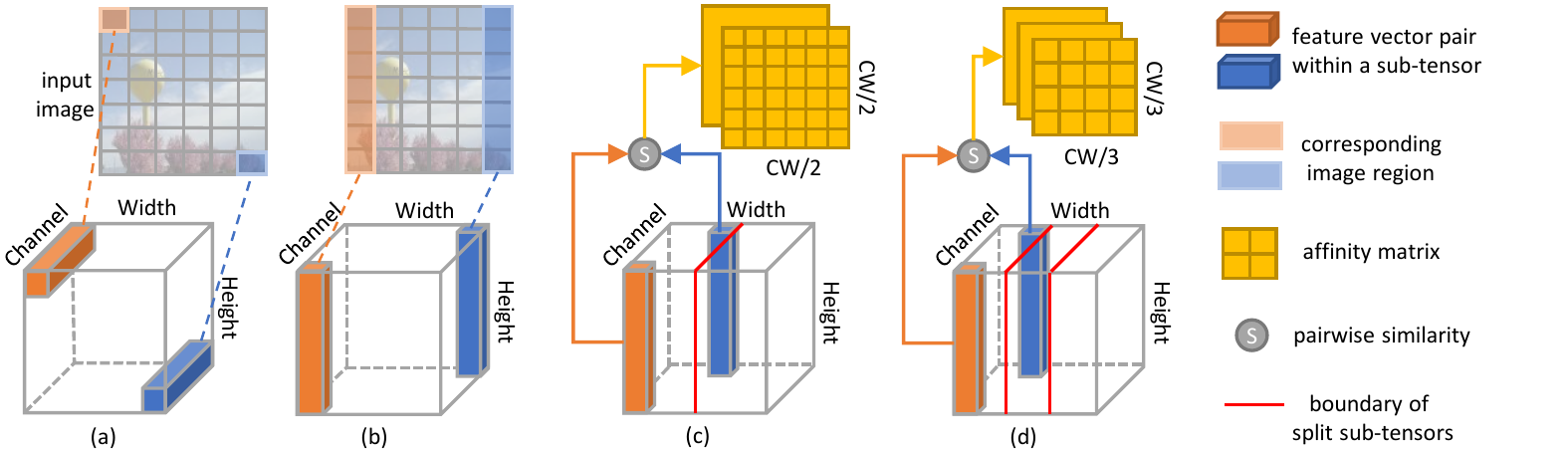}
\caption{(a) Vanilla (Spatial) Non-local Module; (b) Vanilla Depthwise Non-local Module (vertical-split); (c) Divide-and-Conquer Depthwise Non-local Module (vertical-split, split number, $s$=2); (d) split number, $s$=3. Dense pairwise similarity is computed between each pair of feature vectors (the orange bar and the blue bar) within the same region/sub-region. In (c), the $C \times W$ plane is split into 2 sub-regions.}
\label{fig:divide}
\end{figure*}

To efficiently harvest intra-channel correlations, the depthwise non-local module considers $I_{k,i}$ or $I_{k,j}$ (shown in Fig.~\ref{fig:divide}(b) ) rather than $I_{i,j}$ (shown in Fig.~\ref{fig:divide}(a) ) as a feature vector. $I_{k,i}$ and $I_{k,j}$ represent some feature of their corresponding horizontal and vertical image region respectively. The proposed module is a residual block with two possible types of residual layers. One type of layers is called vertical or \textit{vertical-split} layer and the other is horizontal or \textit{horizontal-split} layer. In a vertical-split layer, we take $I_{k,j}$ as a feature vector. In a horizontal-split layer, $I_{k,i}$ is taken as a feature vector. These two types of layers are designed in a similar and symmetric way.

In the vertical-split layer shown in Fig.~\ref{fig:DNLmodule}, the input feature map can be seen as $C \times W$ feature vectors and each vector has $H$ elements. To exploit cross-channel features, we first compute an attention map by measuring pairwise similarities among these $CW$ vectors. Thus the attention map $A$ is of size $CW \times CW$. Consider two arbitrary features, $I_{k,j}$ and $I_{k',j'}$, whose indices in the attention map are $p$ and $q$. Their attention score can be calculated in a bilinear form.
\begin{equation}
A_{p,q} = I_{k,j}^T U_\theta^T V_\phi I_{k',j'},
\end{equation}
where $U_\theta$ and $V_\phi$ are two learnable matrices of two embedding layers $\theta$ and $\phi$. The size of $U_\theta$ and $V_\phi$ is $H^A \times H$. The bilinear form can be seen as mapping $I_{k,j}$ and $I_{k',j'}$ into a latent space $R^{H^A}$ by linear transformations, and then computing the dot product between the transformed features as similarity. Importantly, we discuss whether all $C \times W$ feature vectors share the same transform matrix $U_\theta$ in the following. $I_{k,j}$ with the same $k$ but different $j$, which denotes different spatial region within the same latent space, can share the same transform matrix. $I_{k,j}$ with different $k$ should not utilize the same transform matrix, since different $k$ implies different latent space.
Thus, only feature vectors with the same channel share the same parameter $U_\theta$. Since the layers $\theta$ and $\phi$ embed different channels of an input feature with different weights, we name $\theta$ and $\phi$ as `Channel-wise Embedding Layer' as displayed in Fig.~\ref{fig:DNLmodule}.
We reformulate the bilinear form (shown as `pairwise correlation' in Fig.~\ref{fig:DNLmodule}) in the below.
\begin{equation}
A_{p,q} = I_{k,j}^T W_{\theta,k}^T W_{\phi,k'} I_{k',j'}
\end{equation}
where $W_{\theta}$ and $W_{\phi}$ are $C\times H^A\times H$ matrices. $W_{\theta,k}$ and $W_{\phi,k'}$ are $H^A\times H$ matrices corresponding to $U_\theta$ and $V_\phi$.
We claim that the above attention model is promising in modeling contrast and connectivity for salient object detection. Small $A_{p,q}$ indicates high contrast between $I_{k,j}$ and $I_{k,j'}$. If $I_{k,j}$ and $I_{k,j'}$ are spatially close to each other, large $A_{p,q}$ suggests that their corresponding regions are connected in the saliency map.

To leverage these cues, we conceive a strategy to propagate contrast information or saliency value from $I_{k',j'}$ to $I_{k,j}$, according to their correlation. The strategy learns a feature residual by taking attention weighted transformation of the feature map.
\begin{equation}
\tilde{I} = f(\mbox{softmax}(A) g(I)),
\end{equation}
where $\tilde{I}$ denotes the above-mentioned feature residual. $\mbox{softmax}(\cdot)$ normalizes each row of matrix $A$. $g(\cdot)$ denotes a transformation of the input feature map $I$. As defined below, $g(\cdot)$ adopts linear transformations with different parameters for different channels of $I$. Let $g(I) = [ I_1^TW_{g,1}, ... I_k^TW_{g,k}, ... I_C^TW_{g,C}]$,
where $I_k$ is a $H\times W$ matrix representing the $k$-th channel of feature map $I$, $W_{g,k}$ is a $H\times H'$ matrix representing the linear transformation for the $k$-th channel of the feature map, $[\cdot]$ denotes the concatenation along the first dimension. Thus $g(I)$ is a $CW\times H'$ matrix. $\mbox{softmax}(A)g(I)$ is an attention weighted linear transformation of $g(I)$, shown as `Matrix Multiplication' in Fig.~\ref{fig:DNLmodule}. Since the size of $\mbox{softmax}(A)g(I)$ is $CW\times H'$, another transformation $f$ is required to map it into the $R^H$ space.
Let $y = \mbox{softmax}(A)g(I)$. Then $f(y) = [(y_1 W_{f,1})^T, ...(y_k W_{f,k})^T, ... (y_C W_{f,C})^T]$,
where $y$ is a $CW\times H'$ matrix, and $y$ is reshaped into a $C\times W\times H'$ tensor before $f(\cdot)$ is applied, $W_f$ is a $C\times H'\times H$ tensor, and $f(y)$ is a $CH\times W$ matrix. Finally, $f(y)$ is reshaped into a $C\times H\times W$ residual tensor $\tilde{I}$, and $I$ is updated by adding the residual tensor and $I$ together. The output of DNL modules is calculated as $O = I + \tilde{I} = I + f(y)$, where $O$ represents the output feature map, and the reshaping operators are omitted.

The horizontal-split layer is a symmetric form of the vertical-split layer. We simply summarize its process and describe the differences from the vertical one.
\begin{eqnarray}
    A_{p,q} = I_{k,i}^T W_{\theta,k}^T W_{\phi,k'} I_{k',i'}, \\
    g(I) = [I_1 W_{g,1}, ... I_k W_{g,k}, ... I_C W_{g,C}], \\
    O = I + [y_1 W_{f,1}, ...y_k W_{f,k}, ...y_C W_{f,C}],
\end{eqnarray}
where the size of attention map $A$ is $CH\times CH$, and $A_{p,q}$ denotes pairwise similarity between two arbitrary features $I_{k,i}$ and $I_{k',i'}$ ($1\leq k, k'\leq C$, $1\leq i, i'\leq H$), whose indices in the attention map $A$ are $p$ and $q$. $I_{k,i}$ is a feature vector of length $W$. $W_\theta$, $W_\phi$, $W_g$ and $W_f$ are $C\times W^A\times W$, $C\times W^A\times W$, $C\times W\times W'$ and $C\times W'\times W$ tensors, respectively. $y$ is computed in the same way as in the vertical-split layer, and is reshaped into a $C\times H\times W'$ tensor before $f$ is applied. $f(y)$ is converted to a $C\times H\times W$ tensor before $I$ is updated. Note that all channel-wise embedding layers, $\theta$, $\phi$, $g(\cdot)$ and $f(\cdot)$, have bias parameters. For example, $y_k W_{f,k}$ should actually be $[y_k^{*T}, 1]^T[W_{f,k}^*, B_{f,k}]$. For simplicity, all bias terms have been omitted in the above formulations.

\subsection{Divide-and-Conquer}\label{sec:divide-conquer}
In this section, we accelerate the naive depthwise non-local module by dividing an input feature map into multiple sub-tensors. A few rationales support the divide-and-conquer strategy. First, the naive DNL module computes dense pairwise similarities, which is too computationally expensive for a fast neural network module. Second, the divide-and-conquer strategy still maintains spatial coherence in the resulting saliency map. If there is a strong similarity between spatially adjacent features, it is most likely that these two features or segments belong to the same object. Propagating saliency values between such pairs of features can likely improve the accuracy of saliency prediction. In the naive vertical-split DNL module, all pairs of feature vectors on the $C \times W$ plane are used to measure similarity, as shown in Fig.~\ref{fig:divide}(b). For the divide-and-conquer DNL shown in Fig.~\ref{fig:divide}(c), the feature tensor is divided into 2 sub-tensors, and vector pairs are only sampled from the same sub-tensor. For each sub-tensor, a smaller affinity matrix is obtained by calculating the pairwise correlation. The softmax operation is separately applied for each affinity matrix. Different sub-tensors still share the same $W_\theta$ and $W_\phi$. The number of sub-tensors is controlled by split number $s$.

\subsection{Complexity Analysis}
In this section, we analyze the space and time complexities of the vanilla depthwise non-local module and its divide-and-conquer version. This analysis can help us determine the values of hyper-parameters and the location of the proposed module in our network architecture.

Let us first discuss the space complexity of a vanilla depthwise non-local module. We assume that all variables are released after inference. Only parameters and intermediate variables are considered. The size of $W_\theta$, $W_\phi$, $W_g$ and $W_f$ in a vertical-split layer is respectively $C\times H^A\times H$, $C\times H^A\times H$, $C\times H\times H'$ and $C\times H'\times H$ while their size is $C\times W^A\times W$, $C\times W^A\times W$, $C\times W\times W'$ and $C\times W'\times W$ in a horizontal-split layer. 
Without taking bias terms into account, the total number of parameters is $2C(H(H^A+H')+W(W^A+W'))$.
The size of intermediate variables $A$, $g(I)$, $y$ and $\tilde{I}$ is respectively $CW\times CW$, $CW\times W'$, $CW\times W'$, and $C\times H\times W$ in a vertical layer. In a horizontal one, their size is $CH\times CH$, $CH\times H'$, $CH\times H'$, and $C\times H\times W$. The space complexity of intermediate variables is $C^2(H^2+W^2)+2C(HH'+HW+WW')$. The space complexity of a depthwise non-local module is $\mathcal{O}(C^2(H^2+W^2))$.

For time complexity, we count the number of multiplications and additions (MAdds). In a vertical-split layer, applying transformations $W_\theta$ and $W_\phi$ costs $CWHH^A$ while computing pairwise similarity costs $C^2W^2H^A$. The time complexity of $\mbox{softmax}(\cdot)$, $g(\cdot)$, $f(\cdot)$ and $\mbox{softmax}(A)g(I)$ is respectively $C^2W^2$, $CWHH'$, $CWHH'$ and $C^2W^2H'$. In a horizontal-split layer, computing $A$ costs $2CHWW^A + C^2H^2W^A$. The time complexity of $\mbox{softmax}(\cdot)$, $g(\cdot)$, $f(\cdot)$ and $\mbox{softmax}(A)g(I)$ is respectively $C^2H^2$, $CHWW'$, $CHWW'$ and $C^2H^2W'$. The total number of multiplications and additions is $C^2W^2(H^A+H'+1) + C^2H^2(W^A+W'+1) + 2CHW(H^A+H'+W^A+W')$. The time complexity of the proposed module is $\mathcal{O}(C^2HW(H+W))$.

Next, we analyze the computational cost of a divide-and-conquer depthwise non-local module. Its space complexity is reduced by a factor of $s$ since the size of $A$ becomes $\frac{1}{s}C^2W^2$ in a vertical layer and $\frac{1}{s}C^2H^2$ in a horizontal layer. As for time complexity, computing attention scores in a vertical-split layer costs $2CWHH^A + s (\frac{CW}{s})^2 H^A$. The time complexity of $\mbox{softmax}(\cdot)$, $g(\cdot)$, $f(\cdot)$ and $\mbox{softmax}(A)g(I)$ is respectively $\frac{C^2W^2}{s}$, $CWHH'$, $CWHH'$ and $s(\frac{CW}{s})^2H'$. The computation in a vertical-split layer costs $\frac{1}{s}C^2W^2(H^A+H'+1)+2CHW(H^A+H')$ and a horizontal-split layer $\frac{1}{s}C^2H^2(W^A+W'+1)+2CHW(W^A+W')$. The time complexity of the accelerated variant is $\mathcal{O}(\frac{1}{s}C^2HW(H+W))$. Notice that $H^A$/H' and $W^A$/W' are set as H/2 and W/2 respectively in our implementation.
\begin{table}[!t]
	\caption{Complexity of Vanilla Depthwise Non-Local Module and its Divide-and-Conquer Variant.}
	\centering
	\resizebox{\linewidth}{!}{
		\begin{tabular}{lccc}
			\toprule
			Complexity & Vanilla                     & Divide-and-Conquer  \\
			\midrule
			Space      & $\mathcal{O}(C^2(H^2+W^2))$ & $\mathcal{O}(\frac{1}{s}C^2(H^2+W^2))$ \\
			Time       & $\mathcal{O}(C^2HW(H+W))$   & $\mathcal{O}(\frac{1}{s}C^2HW(H+W))$ \\
			\bottomrule
		\end{tabular}
	}
	\label{table:complexity}
\end{table}

\begin{figure}[h]
\centering
\includegraphics[width=1.0\linewidth]{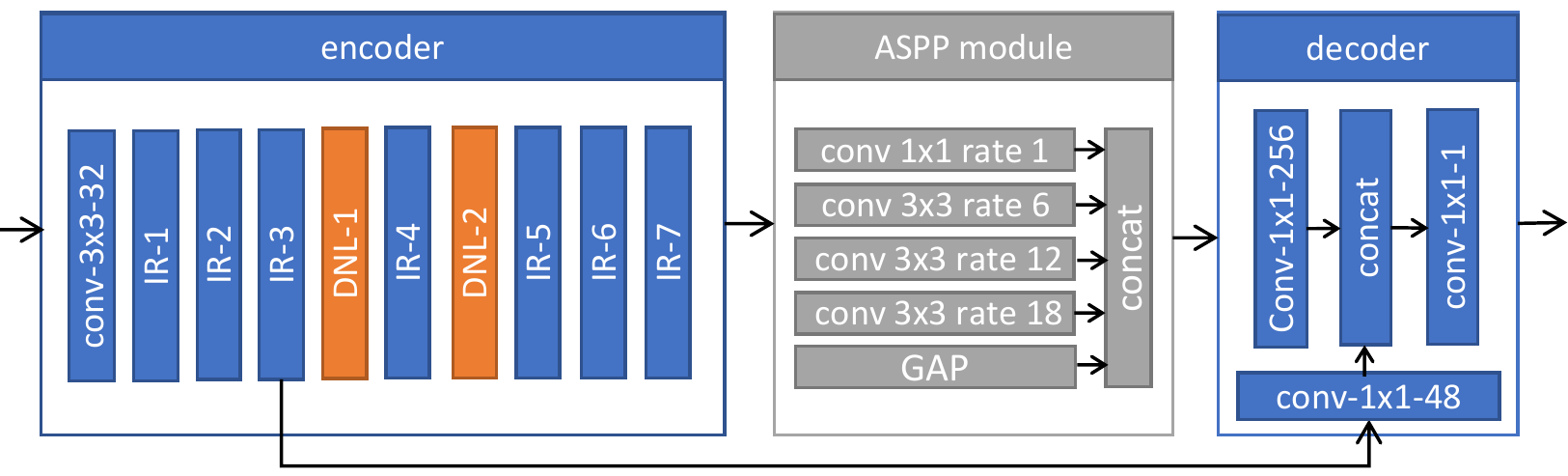}
\caption{Depthwise Non-Local Deep Neural Network. As shown in the above, IR denotes inverted residual module while DNL denotes depthwise non-local module. `rate' refers to the dilation rate of a convolution. GAP denotes global average pooling.}
\label{fig:DNLnet}
\end{figure}
\subsection{Model Architecture}
In this section, we develop a depthwise non-local neural network based on our proposed module. The proposed network consists of an encoder, an atrous spatial pyramid pooling (ASPP) module~\cite{Chen_2018_ECCV} and a decoder, as shown in Figure~\ref{fig:DNLnet}. The encoder contains 7 inverted residual modules, 2 depthwise non-local modules and several regular convolution layers. Each inverted residual (IR) module is composed of one or multiple inverted residual blocks. The hyper-parameters of these above-mentioned IR modules follow the setting in~\cite{sandler2018mobilenetv2}.

Our proposed depthwise non-local modules are located between some of the inverted residual modules to strengthen non-local correlations among all feature map channels. Since the computational complexity of the proposed module grows quadratically with respect to the number of channels and cubically w.r.t the spatial resolution of input feature map, we place the proposed module at middle levels of the network. The first depthwise non-local module is placed behind the third inverted residual module. The second one is inserted after the fourth inverted residual module. After the third inverted residual module, the feature map has shrunk to the smallest spatial size, $1/8$ of the original image size, while the number of channels is still reasonably small. Thus positioning the proposed modules at middle levels helps lower the computational cost incurred by these modules. Theoretically DNL modules can be placed at any position of a backbone network. How the position of a DNL module affect its performance and efficiency is investigated in Section~\ref{sec:abla_study}.

The atrous spatial pyramid pooling module in Fig.~\ref{fig:DNLnet} concatenates the feature maps produced from the five parallel layers along the depth dimension. These five parallel layers are respectively a $1\times 1$ pointwise convolution, three dilated $3\times 3$ convolutions and an image pooling layer. The dilation rates of the three atrous convolutions are 6, 12, and 18 respectively. All of these five parallel layers produce 256-channel feature maps. The image pooling layer consists of a global spatial averaging sub-layer and a pointwise convolution that converts the number of output channels to 256. The ASPP module produces a 1280-channel feature map. The decoder takes the output of both the ASPP module and the third inverted residual module as a combination of high-level and low-level inputs. The decoder reduces the number of channels in the high-level input to 256 while increasing the number of channels in the low-level input to 48 using pointwise convolutions following a 2D batch normalization layer. Finally, the low-level and high-level features are concatenated to predict a dense saliency map via a $1\times 1$ convolution.

\section{Experiments}
In the experiment section, salient object detection methods are tested on DUT-OMRON~\cite{yang2013saliency}, ECSSD~\cite{yan2013hierarchical}, HKU-IS~\cite{li2016visual} test set, PASCAL-S~\cite{li2014secrets} and DUTS~\cite{wang2017learning}. All the above datasets provide dense pixel-level annotations. DUT-OMRON contains 5168 challenging images which has one or more salient objects. ECSSD has 1000 images. HKU-IS includes a train set of 2500 images, a validation set of 500 images and a test set of 1447 images. PASCAL-S consists of 850 images. Threshold is chosen as 0.5 to binarize masks of PASCAL-S, as suggested in~\cite{li2014secrets}. Notice that all salient object detection models are not trained on any subsets of DUT-OMRON and all 5168 images are utilized as testing samples. Thus DUT-OMRON is a challenging benchmark which can reveal the generalization capability of a salient object detection model. HKU-IS is another challenging dataset in which many images contain multiple ground-truth objects. Our proposed method is trained with 5000 images from MSRA-B train set and HKU-IS train set. The optimization algorithm is SGD with weight decay 5e-4, momentum 0.9 and initial learning rate 1e-7. We adopt poly policy with power 0.9 to tune the learning rate. The proposed network is trained for 300 epochs. Pytorch 0.4.1 with MKL backend is used for all deep learning methods. For the methods whose released model is not trained with Pytorch, their model weights are copied to an implementation of Pytorch. For fair comparisons, the efficiency of the mentioned salient object detection algorithms are evaluated on the same personal computer, which has an Intel i7-6850k CPU with 3.60 GHz base frequency, a GeForce GTX 1080 Ti GPU and 32GB memory.
\begin{table*}[!ht]
	\caption{Comparison among the state-of-the-art and ours. $*$ methods are originally proposed for image classifications and segmentations.}
	\centering
	\normalsize
	\resizebox{\linewidth}{!}{
		\begin{tabular}{lrrrrrrrrrrrrrrr}
			\toprule
			& \multicolumn{3}{c}{DUT-OMRON} & \multicolumn{3}{c}{ECSSD} & \multicolumn{3}{c}{HKU-IS} & \multicolumn{3}{c}{PASCAL-S} & \multicolumn{3}{c}{DUTS}\\
			& maxF     & MAE      & S-m      & maxF    & MAE     & S-m    & maxF    & MAE     & S-m     & maxF     & MAE      & S-m     & maxF    & MAE      & S-m \\
			\midrule
			RAS    & \color{red}{0.7848} & \color{red}{0.0633} & \color{red}{0.8119} & \mygreen{0.9203} & \color{red}{0.0551} & \color{red}{0.8935} & 0.9116 & \color{red}{0.0449} & 0.8875 & \mygreen{0.8319} & 0.1021 & 0.7978 & \mygreen{0.8310} & \mygreen{0.0591} & \color{red}{0.8281} \\
			PAGRN  & 0.7709 & \mygreen{0.0709}   & 0.7751   & \color{red}{0.9268}  & \mygreen{0.0609}  & \mygreen{0.8892} & \color{red}{0.9187}  & 0.0475  & \mygreen{0.8891}  & \color{red}{0.8531}   & \color{red}{0.0921}   & \color{red}{0.8190}  & \color{red}{0.8541} & \color{red}{0.0549} & \mygreen{0.8254} \\
			UCF    & 0.7365   & 0.1318   & 0.7578   & 0.9097  & 0.0790  & 0.8816 & 0.8866  & 0.0749  & 0.8643  & 0.8217   & 0.1292   & 0.7999  & 0.7700 & 0.1178 & 0.7710 \\
			NLDF   & 0.7532   & 0.0796   & 0.7704   & 0.9050  & 0.0626  & 0.8747 & 0.9017  & 0.0480  & 0.8782  & 0.8278   & \mygreen{0.0990}   & \mygreen{0.8036}  & 0.8156 & 0.0649 & 0.8052 \\
			DSS    & 0.7604   & 0.0744   & 0.7892   & 0.9078  & 0.0620  & 0.8836 & 0.9005  & 0.0499  & 0.8805  & 0.8262   & 0.1029   & 0.8025 & 0.8130 & 0.0647 & 0.8168 \\
			RFCN   & 0.7332   & 0.0782   & 0.7503   & 0.8867  & 0.0765  & 0.8368 & 0.8832  & 0.0572  & 0.8361  & 0.8284 & 0.0996   & 0.7895  & 0.7702 & 0.0744 & 0.7506 \\
			DCL    & 0.7260   & 0.0944   & 0.7498   & 0.8884  & 0.0717  & 0.8672 & 0.8823  & 0.0584  & 0.8650  & 0.8053   & 0.1092   & 0.7930 & 0.7756 & 0.0787 & 0.7859 \\
			DS     & 0.7449   & 0.1204   & 0.7502   & 0.8824  & 0.1217  & 0.8206 & 0.8661  & 0.0791  & 0.8531  & 0.8109   & 0.1472   & 0.7715 & 0.7756 & 0.0894 & 0.7916 \\
			MDF    & 0.6944   & 0.0916   & 0.7208   & 0.8316  & 0.1050  & 0.7761 & 0.8605  & 0.1291  & 0.8101  & 0.7655   & 0.1451   & 0.6935  & 0.7285  & 0.0995 & 0.7232 \\
			\midrule
			$*$MoblieNetV2 & 0.7446 & 0.0871 & 0.7595 & 0.8901 & 0.0737 & 0.8614 & 0.8979 & 0.0537 & 0.8756 & 0.7716 & 0.1318 & 0.7592 & 0.7613 & 0.0806 & 0.7720 \\
			$*$ShuffleNet    & 0.7300 & 0.0946 & 0.7506 & 0.8763 & 0.0830 & 0.8443 & 0.8914 & 0.0558 & 0.8701 & 0.7856 & 0.1216 & 0.7709 & 0.7698 & 0.0845 & 0.7740 \\
			\midrule
			ours   & \mygreen{0.7795}   & 0.0779   & \mygreen{0.7981}   & 0.9096  & 0.0646  & 0.8851 & \mygreen{0.9133}  & \mygreen{0.0451}  & \color{red}{0.8974}  & 0.8229   & 0.1065   & 0.8031 & 0.8081 & 0.0731 & 0.8071 \\
			\bottomrule
		\end{tabular}
	}
	\label{table:comp1}
\end{table*}

\subsection{Comparison on Quality of Saliency Maps}
To evaluate the quality of saliency maps, we adopt maximum F-measure (maxF)~\cite{achanta2009frequency}, mean absolute error (MAE) and structure-measure~\cite{fan2017structure} (S-m) as criteria. To compute maximum F-measure, we first sample a list of thresholds. Given a threshold, the average of precision and recall is computed for all saliency predictions in a dataset. Then $F_\beta$ is defined as:
\begin{equation}
F_{\beta} = \frac{(1+\beta^2) \cdot Precision \cdot Recall }{\beta^2 \cdot Precision + Recall}
\end{equation}
where $\beta$ controls the relative importance between precision and recall. $\beta^2$ is selected as 0.3, according to~\cite{achanta2009frequency}. MAE is computed as the average of pixel-level absolute difference between predictions and ground-truth annotations, as shown in:
\begin{equation}
MAE = \frac{1}{HW} \sum_{h=1}^H \sum_{w=1}^W | P_{h,w} - G_{h,w} |
\end{equation}
where $P$ denotes a binarized saliency prediction and $G$ denotes its corresponding binary ground-truth. $H$ and $W$ are height and width of images. $h$ and $w$ are the corresponding indices. Different from estimating pixel-wise errors, S-measure, recently proposed in~\cite{fan2017structure} is adopted to estimate structural similarity between predictions and ground-truth. It is defined as a weighted sum of an object-aware measure and a region-aware measure. Formal definition of structure-measure can be found in ~\cite{fan2017structure}.
\begin{figure*}[!ht]
\centering
\includegraphics[width=1.0\linewidth]{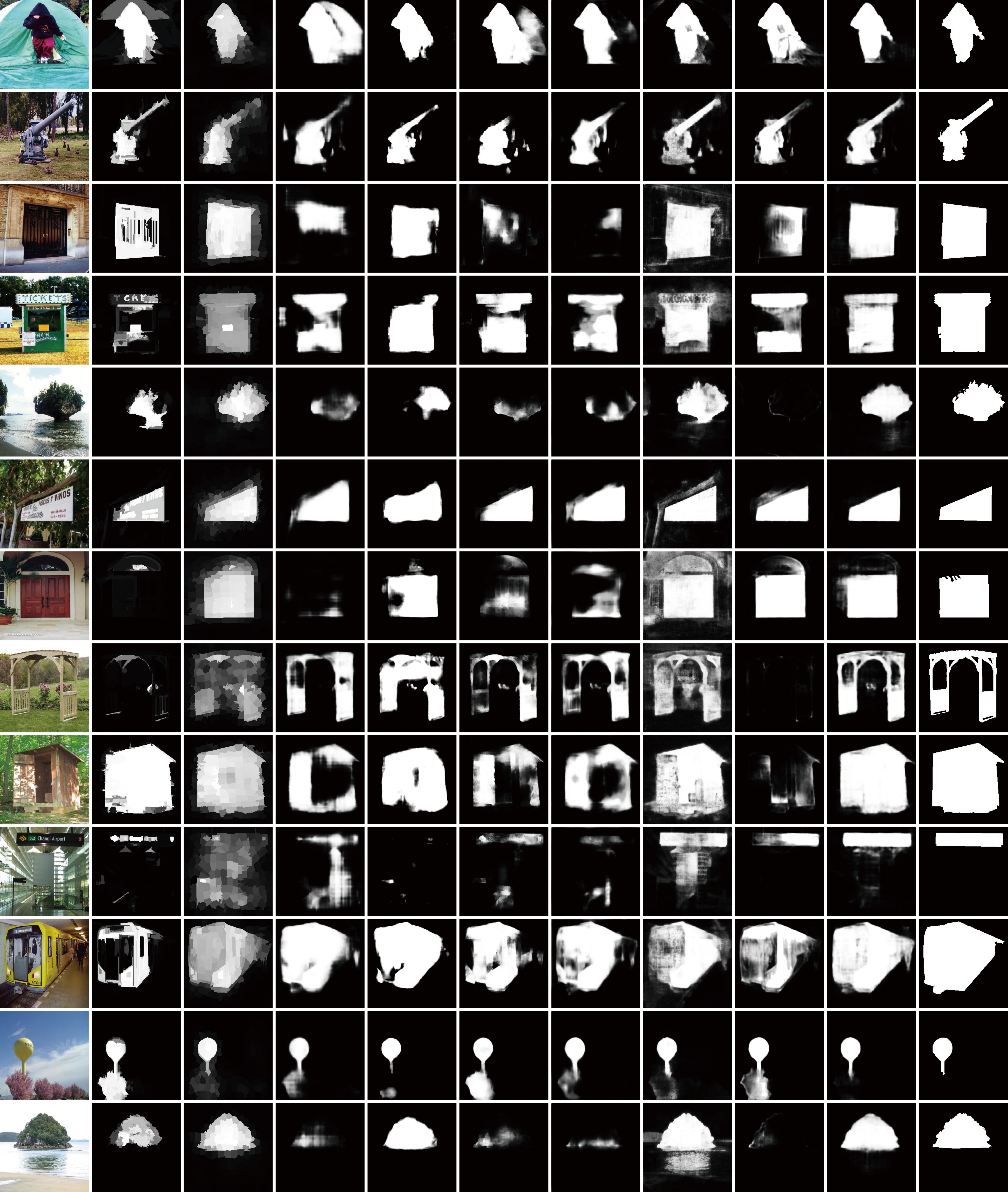}
\leftline{\hspace{0.4cm}input \hspace{0.45cm}MDF~\cite{li2016visual} \hspace{0.25cm}DS~\cite{li2016deepsaliency} \hspace{0.25cm}DCL~\cite{li2016deep} \hspace{0.01cm}RFCN~\cite{wang2016saliency} \hspace{0.01cm}NLDF~\cite{luo2017non} \hspace{0.05cm}DSS~\cite{HouPami18Dss} \hspace{0.15cm}UCF~\cite{zhang2017learning} \hspace{0.1cm}PAGRN~\cite{zhang2018progressive} \hspace{0.35cm}ours \hspace{0.95cm}GT}
   \caption{Qualitative comparison among the state-of-the-art and ours. As shown in the above, the proposed method is compared with MDF, DS, DCL, RFCN, NLDF, DSS, UCF and PAGRN on the DUT-OMRON benchmark. Our proposed method successfully segments complete foreground objects with consistent saliency value and sharp boundaries.}
\label{fig:QualityComp}
\end{figure*}
\begin{figure*}[t]
\centering
\subfigure[Smeasure-CPU time]{
\label{Fig.Tradeoff.sub.1}
\includegraphics[width=0.25\linewidth]{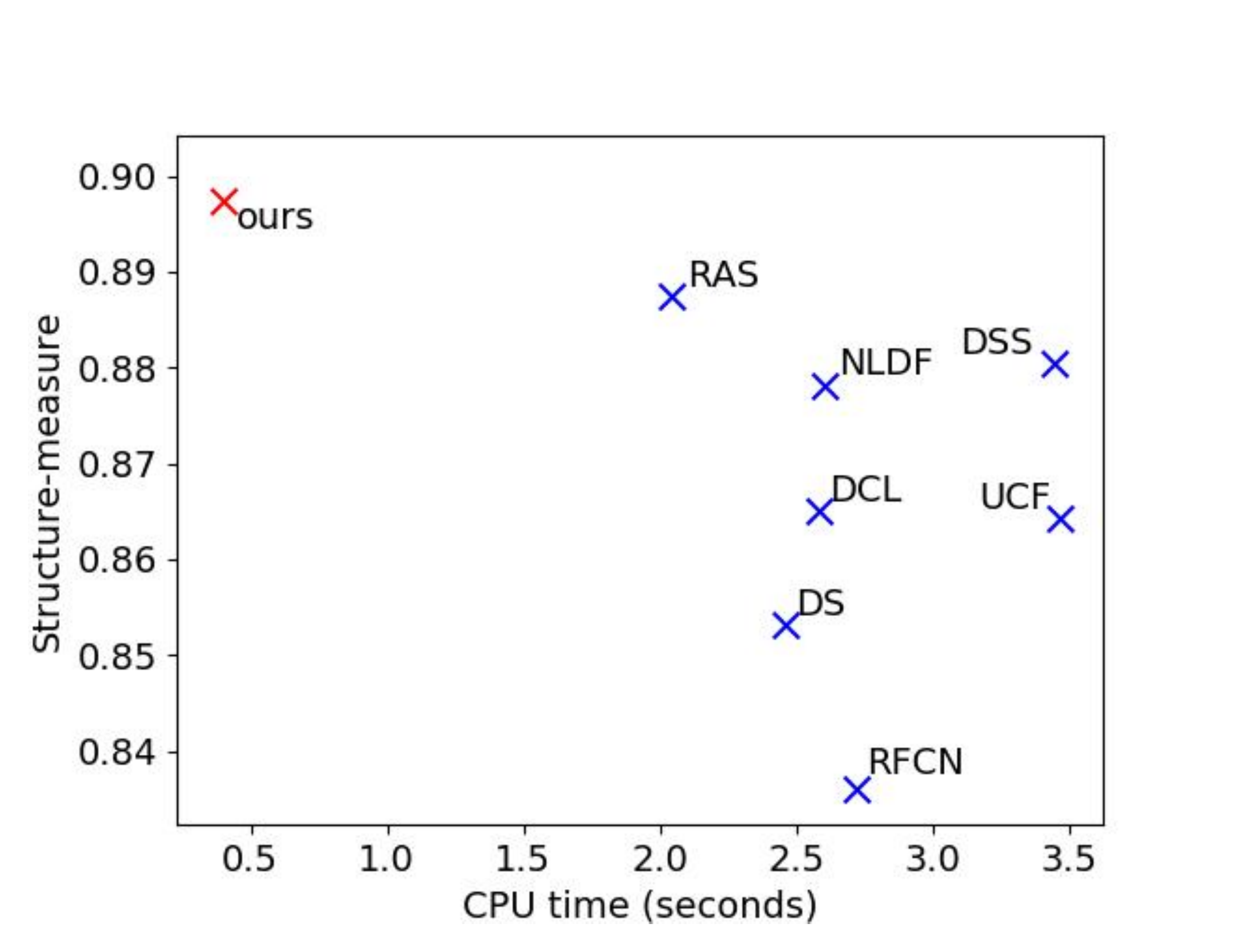}}
\hspace{-0.4cm}
\subfigure[Smeasure-Mem]{
\label{Fig.Tradeoff.sub.2}
\includegraphics[width=0.25\linewidth]{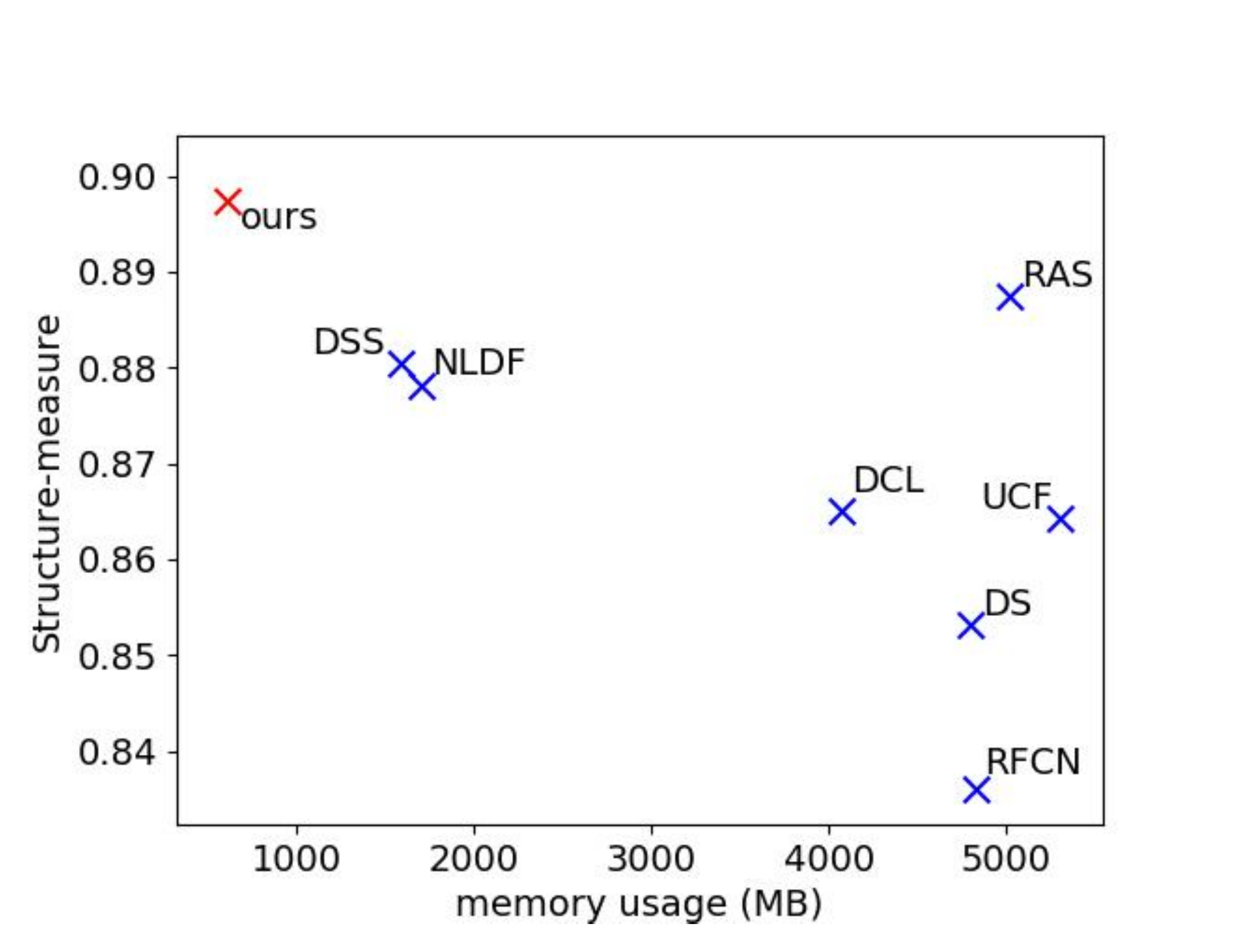}}
\hspace{-0.4cm}
\subfigure[Smeasure-MAdds]{
\label{Fig.Tradeoff.sub.3}
\includegraphics[width=0.25\linewidth]{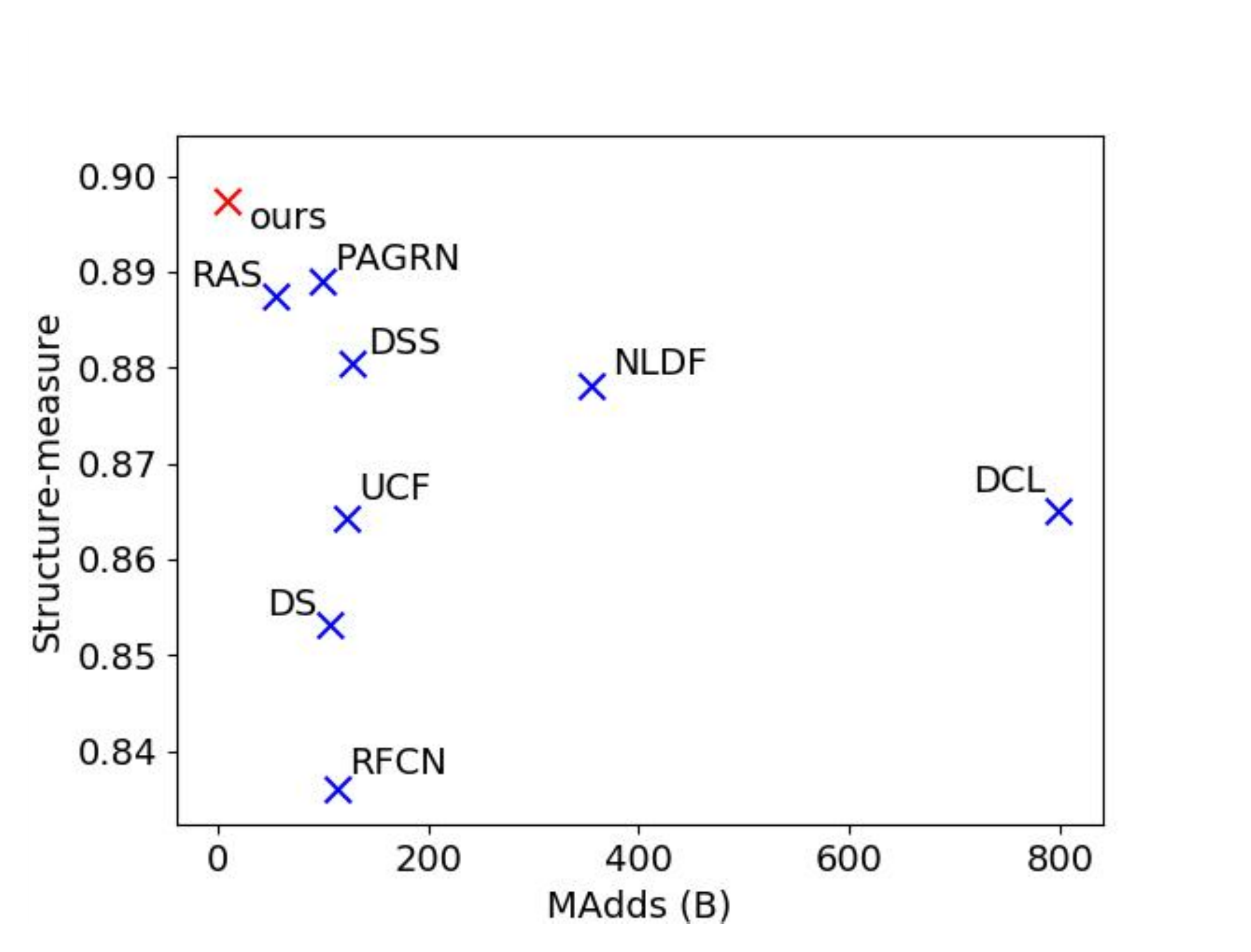}}
\hspace{-0.4cm}
\subfigure[Smeasure-Params]{
\label{Fig.Tradeoff.sub.4}
\includegraphics[width=0.25\linewidth]{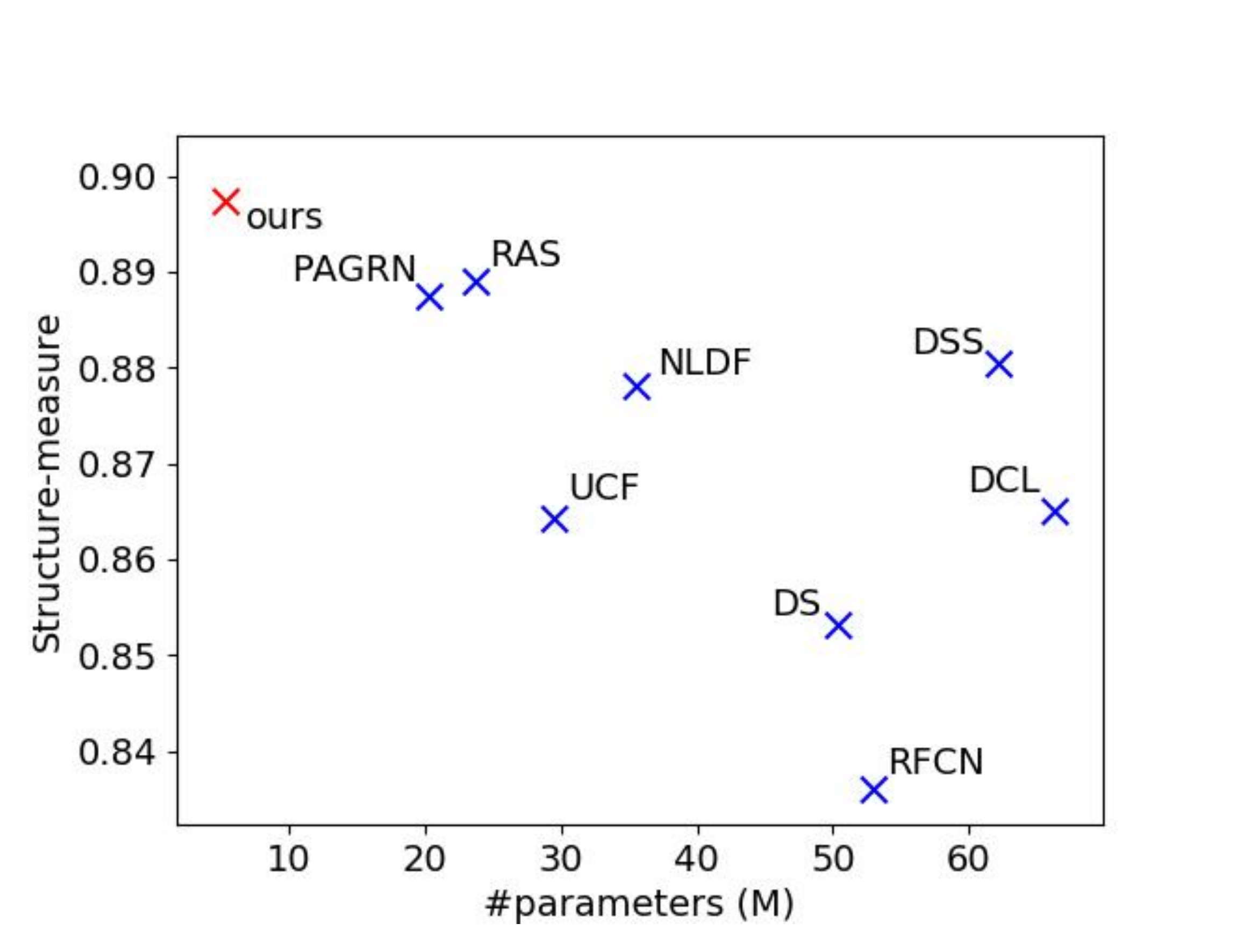}}
\caption{Comparisons on Efficiency and Quality. To measure the efficiency and quality of salient object detection models at the same time, the above scatter diagrams take an efficiency metric as horizontal axis and a quality metric as vertical axis. Our proposed method is always located at the upper left of these diagrams, which indicate the best trade-off between efficiency and accuracy.}
\label{Fig.Tradeoff}
\end{figure*}

As shown in TABLE~\ref{table:comp1} and Fig.~\ref{fig:QualityComp}, the proposed DNL network is compared with existing salient object detection models, RAS~\cite{chen2018reverse}, PAGRN~\cite{zhang2018progressive}, UCF~\cite{zhang2017learning}, NLDF~\cite{luo2017non}, DSS~\cite{HouPami18Dss}, RFCN~\cite{wang2018salient}, DCL~\cite{li2016deep}, DS~\cite{li2016deepsaliency}, MDF~\cite{li2016visual} and two generic lightweight architectures originally proposed for image classification and semantic segmentation, including MobileNet-V2~\cite{sandler2018mobilenetv2} and ShuffleNet~\cite{Zhang_2018_CVPR}.
As suggested in ~\cite{sandler2018mobilenetv2}, MobileNet-V2 serves as an encoder integrated with a DeepLab-V3+ architecture~\cite{Chen_2018_ECCV} to solve semantic segmentation tasks. To detect salient objects, the output channels of the last convolution in DeepLab-V3+ is adjusted to 1. Similar to MobileNet-V2, ShuffleNet is also modified as an encoder with DeepLab-V3+. We fine-tune these two generic frameworks with our training data for saliency prediction. The proposed method significantly outperforms MobileNet-V2 and ShuffleNet on all four benchmarks and three criteria since they fail to capture the contrast as well as the channel-wise coherence information which is essential for saliency inference. Our proposed model obtains the second best maximum F-measure of 0.7795 and the best S-measure of 0.7981 on the large and challenging dataset DUT-OMRON. The DNL network outperforms the third best PAGRN by 0.9\% maxF and DSS by 0.9\% S-measure. Noted that our method is not trained on any subsets of DUT-OMRON but tested on the all 5168 images, which suggests that the DNL network possesses strong generalization capability to achieve stable performance in real applications. The proposed DNL network presents the second best maxF, the second smallest MAE and the best S-measure on the HKU-IS dataset. Particularly, the proposed method surpasses the second best PAGRN by 0.8\% S-measure. Our proposed method show the best results on two challenging benchmarks DUT-OMRON and HKU-IS, which indicates that the proposed network enjoys superior generalization and is comparable to the state-of-the-arts.
\subsection{Comparison on Efficiency}
To evaluate the efficiency of the proposed methods and existing neural network models, this section utilizes CPU time, GPU time, memory usage (denoted as Mem), number of parameters (denoted as Params), MAdds~\cite{sandler2018mobilenetv2} and time complexity as criteria. CPU time is computed using a single CPU thread while GPU time is measured with a single GPU. Batch size is set as 1 for all neural models. Time cost by file input/output is not included but time-consuming preprocessings such as computing prior maps and superpixel segmentation are taken into accounts. Each model sequentially infers 50 randomly selected images from HKU-IS. The peak memory cost during inference is logged as the memory usage. Params is the number of learnable parameters in a neural model and it determines the disk space consumed. MAdds is the number of multiplications and additions, calculated by setting the input size of each method as its default size. Time complexity denoted as `Complexity' in TABLE~\ref{table:comp2} is the number of multiplications and additions with respect to input size that is viewed as variables $H$ and $W$.
\begin{table}[!ht]
\caption{Efficiency of the state-of-the-art and ours.}
\centering
\huge
\resizebox{\linewidth}{!}{
\begin{tabular}{lcccccc}
\toprule
         & CPUTime             & GPUTime                 & Mem               & Params     & MAdds    &  Complexity \\
         &  /secs              & /secs                   & /MB               & /M         & /B       &  /HW \\
\midrule
RAS      & \mygreen{2.0457} & 0.0355 & 5023 & \mygreen{20.23} & \mygreen{54.56} & \mygreen{421.0K} \\
PAGRN    &\rule[1.0mm]{0.8cm}{0.1mm}&\rule[1.0mm]{0.8cm}{0.1mm}&\rule[1.0mm]{0.8cm}{0.1mm}&23.63& 100.4 & 805.8K  \\
UCF      & 3.4696                  & 0.0886                  & 5307                   & 29.43     & 123.3 & 614.4K  \\
NLDF     & 2.6051                  & 0.0340                  & 1709                   & 35.48     & 354.6 & 2863K  \\
DSS      & 3.4451                  & \mygreen{0.0339}   & \mygreen{1587}   & 62.22     & 127.5 & 984.0K  \\
RFCN     & 2.7190                  & 0.0691                  & 4833                   & 53.00     & 113.9 & 455.7K  \\
DCL      & 2.5820                  & 0.0867                  & 4069                   & 66.31     & 797.6 & 3031K  \\
DS       & 2.4588        & 0.0609                  & 4799                   & 50.37     & 106.7 & 426.7K \\
MDF      & 897.68                  & 24.996                  & 1591                   & 75.68     & 7533  & 149.9M  \\
ours     & \color{red}{0.3993}     & \color{red}{0.0113}     & \color{red}{605}      & \color{red}{5.320} & \color{red}{9.567} & \color{red}{73.82K} \\
\bottomrule
\end{tabular}
}
\label{table:comp2}
\end{table}

As shown in TABLE~\ref{table:comp2}, MB denotes million bytes. K, M and B denote thousands, millions and billions respectively. $HW$ represents the product of input height and width. Since the implementation of PAGRN is not available, we only present its theoretical efficiency including parameters, MAdds and time complexity. Most existing CNN based methods predict a saliency map with more than 2.5 seconds on CPU while our proposed network takes less than 0.4 seconds to infer an image. Achieving the fastest CPU inference, our method is $5\times$ faster than the second best RAS. The proposed method also demonstrates the most efficient inference with GPU, and it is $3\times$ faster than the second best DSS. Most methods use 1500-5000 MB memory during inference, which is too expensive for a preprocessing component. Meanwhile our proposed method costs 600 MB running memory, less than 40\% of the second best. Besides, our proposed method has the least parameters, less than 25\% of the second least. Most models consume more than 100 MB storage while our method only costs about 20 MB. Our proposed method obtains the minimum MAdds less than 20\% of the second best. The time complexity of the DNL network is also the lowest and 6 times less than the second lowest RAS. Note that the time complexity of DNL modules actually contains terms with respect to $HW(H+W)$. For convenient comparison we simplify the formula by fixing input size $H\times W$ as default size $360\times 360$. To sum up, our proposed network enjoys the fastest inference speed on both CPU and GPU, consumes the least memory and disk storage, and shows the lowest theoretical complexity, in comparison to existing deep learning models.

To simultaneously evaluate the efficiency and quality of our proposed method, we plot an efficiency metric and a quality metric on the same scatter diagram (shown in Fig.~\ref{Fig.Tradeoff}), with the efficiency metric as horizontal axis and the quality metric as vertical axis. For quality metric, larger S-measure means more accurate predictions while smaller value means lower cost for efficiency metric. Thus the best method balancing accuracy and efficiency should be located at the upper-left in an Efficiency-Quality scatter diagram. As shown in Fig.~\ref{Fig.Tradeoff}, our proposed method achieves the best trade-off between efficiency and quality, on the scatter diagrams of Smeasure-CPU time, Smeasure-Mem, Smeasure-MAdds and Smeasure-Params.

\subsection{Ablation Study}\label{sec:abla_study}
\begin{figure}[t]
\centering
\subfigure[DUT-O]{
\label{Fig.PRcurve.sub.1}
\includegraphics[width=0.33\linewidth]{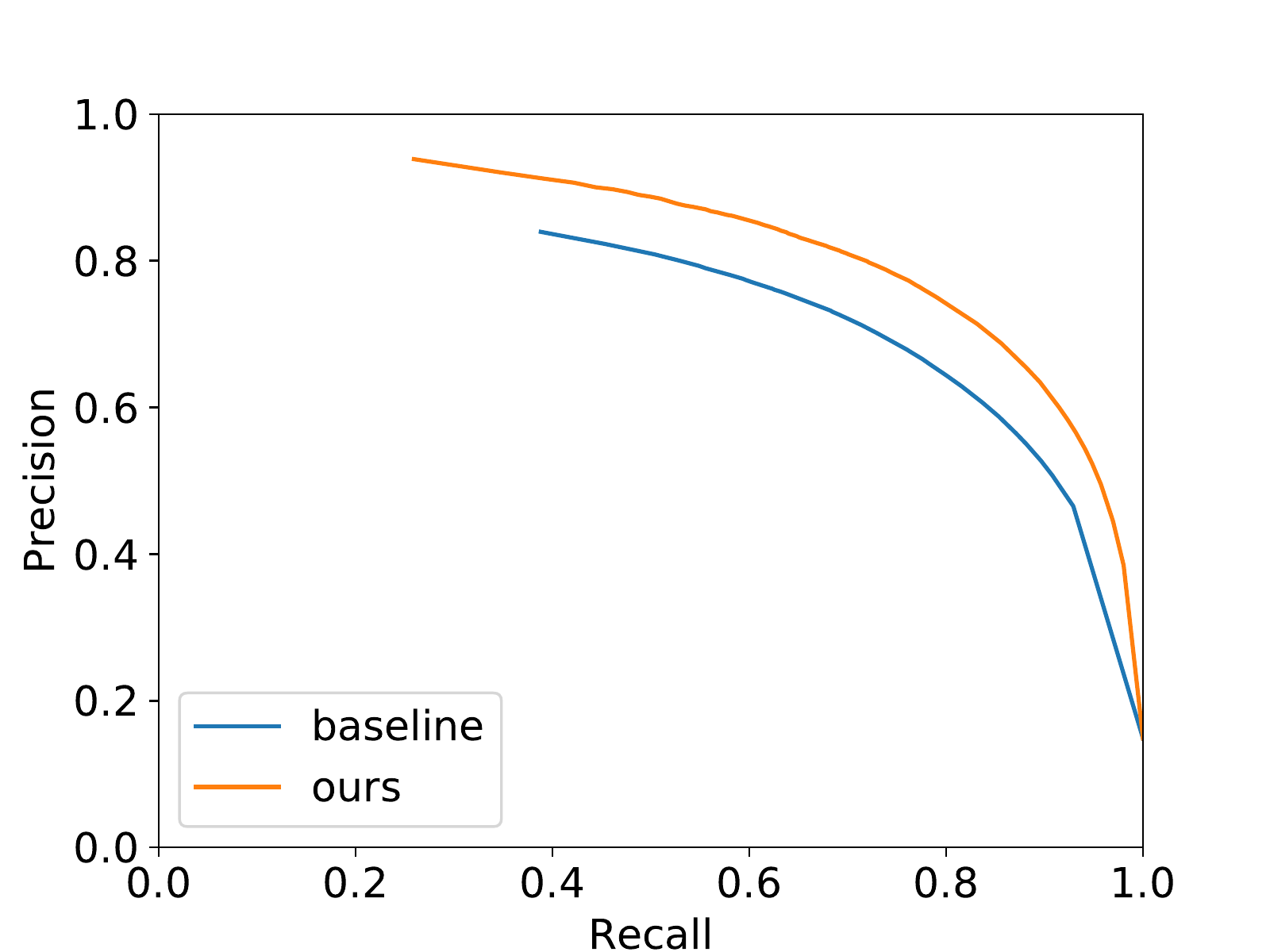}}
\hspace{-0.4cm}
\subfigure[ECSSD]{
\label{Fig.PRcurve.sub.2}
\includegraphics[width=0.33\linewidth]{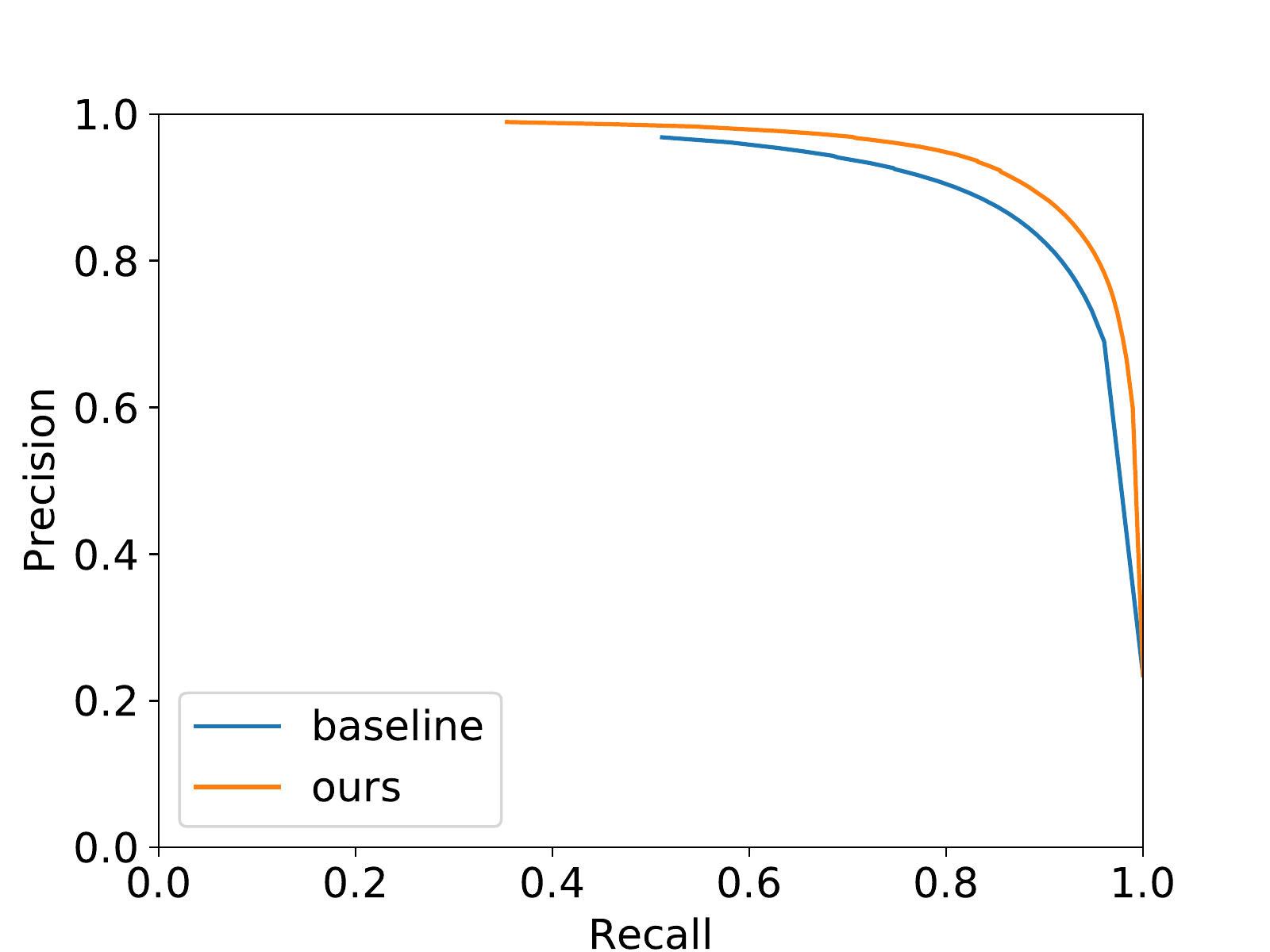}}
\hspace{-0.4cm}
\subfigure[HKU-IS]{
\label{Fig.PRcurve.sub.3}
\includegraphics[width=0.33\linewidth]{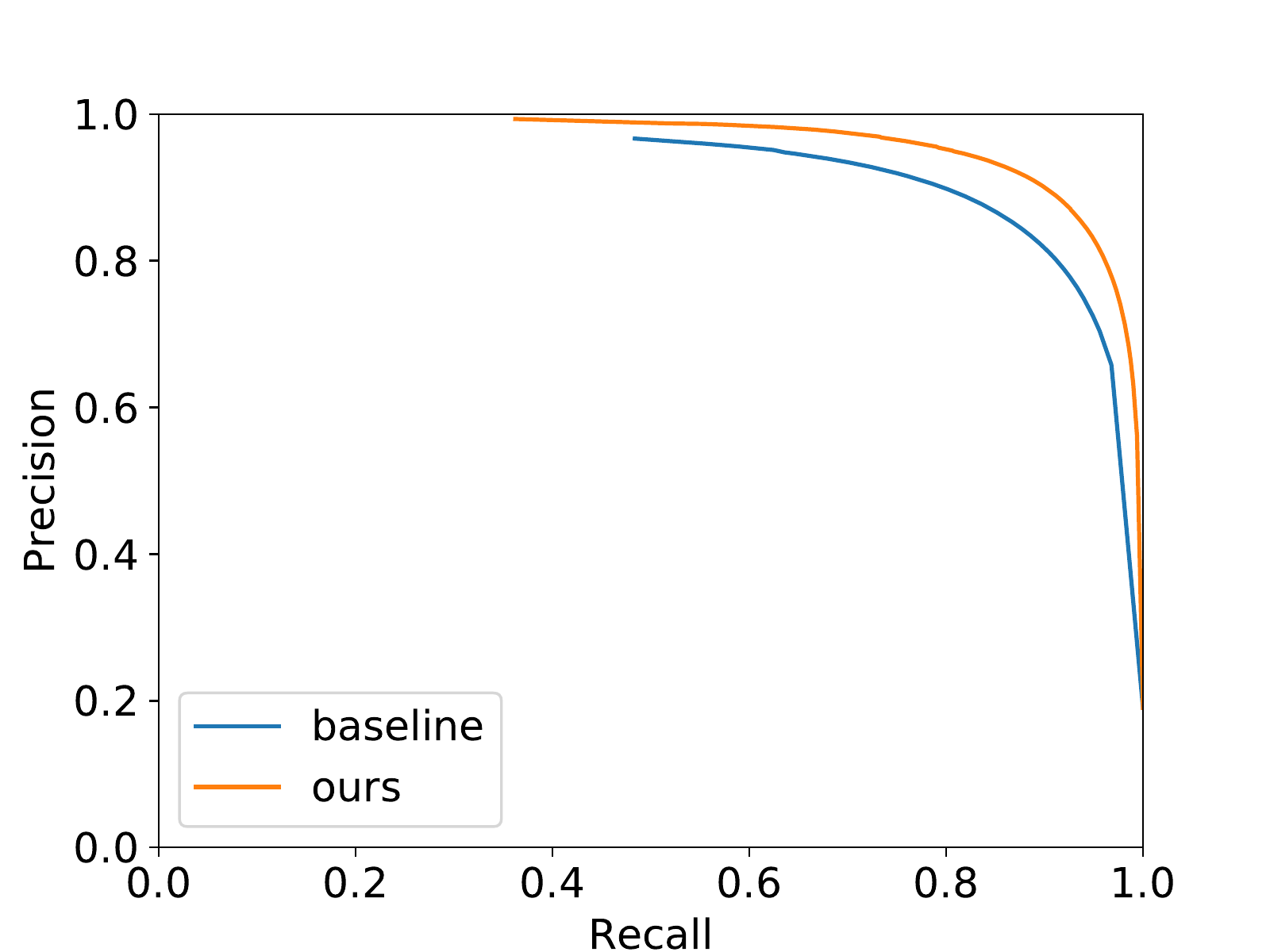}}
\caption{Ablation Study on Precision-Recall Curves. In the above graphs, the blue curve denotes the baseline model without deploying any proposed modules, while the orange one denotes our proposed DNL network. Our proposed method obtains higher precision than the baseline, for the same recall value.}
\label{Fig.PRcurve}
\end{figure}

\begin{figure}[t]
\centering
\subfigure[DUT-O]{
\label{Fig.Fmeasure.sub.1}
\includegraphics[width=0.33\linewidth]{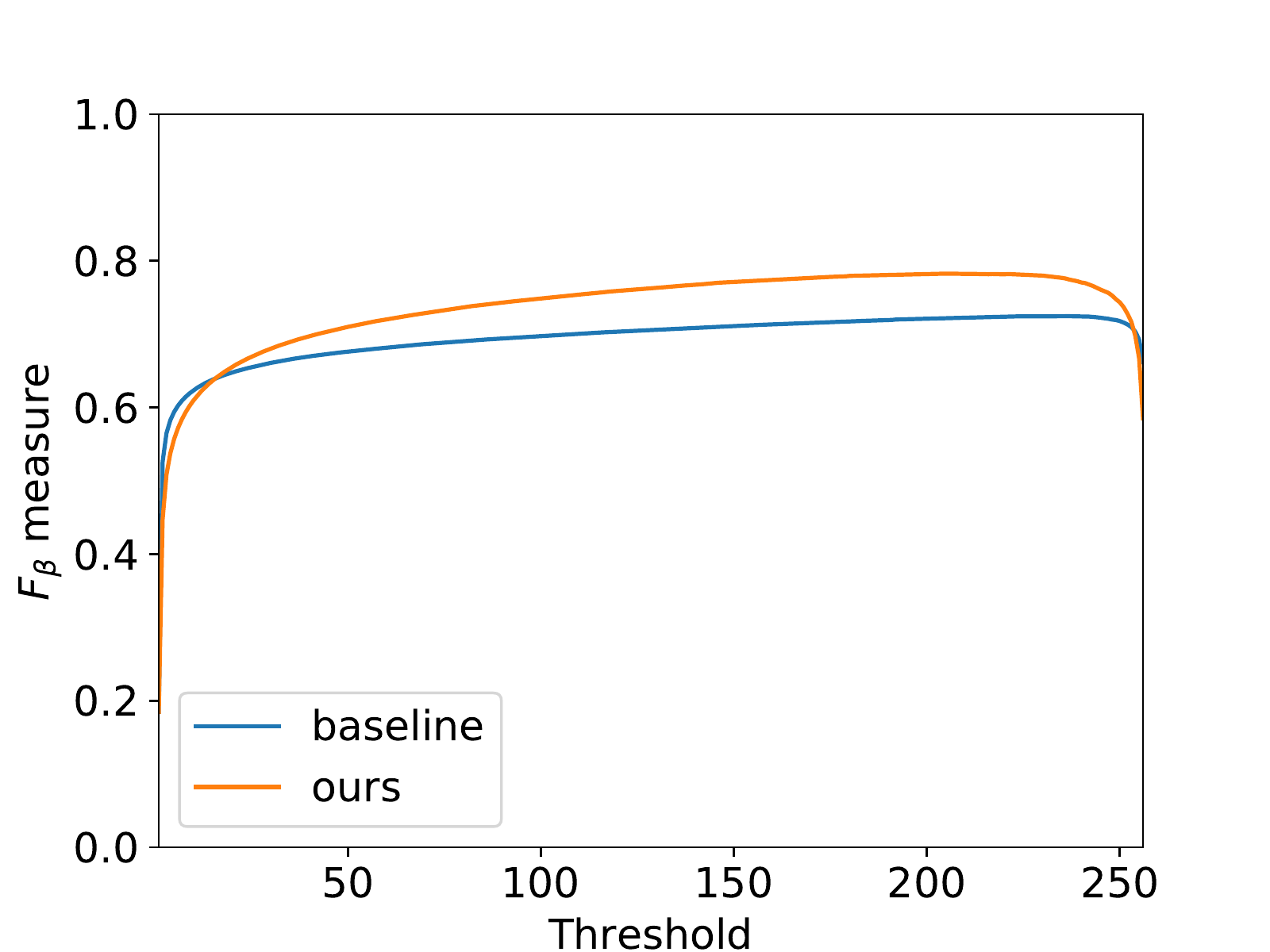}}
\hspace{-0.4cm}
\subfigure[ECSSD]{
\label{Fig.Fmeasure.sub.2}
\includegraphics[width=0.33\linewidth]{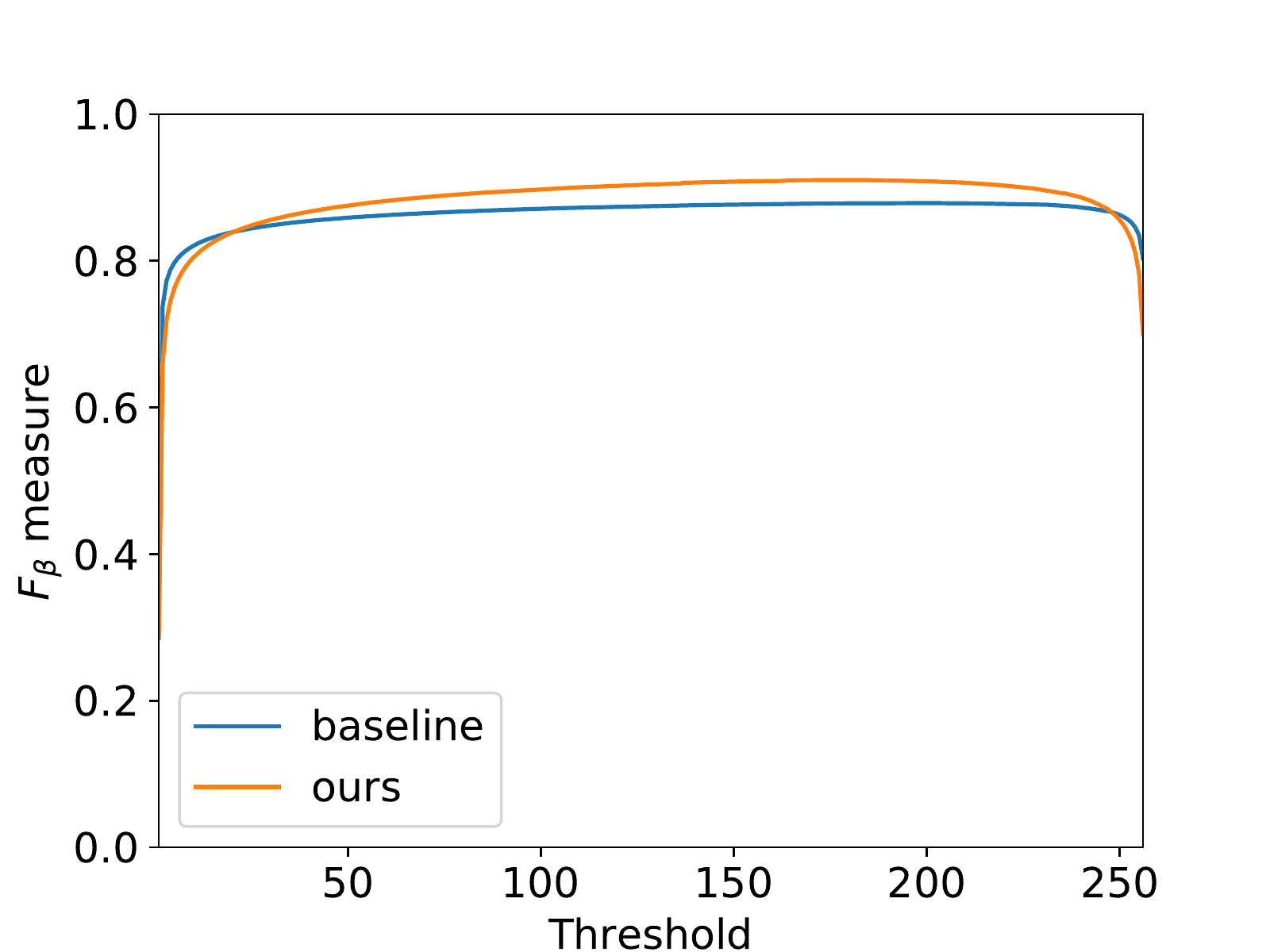}}
\hspace{-0.4cm}
\subfigure[HKU-IS]{
\label{Fig.Fmeasure.sub.3}
\includegraphics[width=0.33\linewidth]{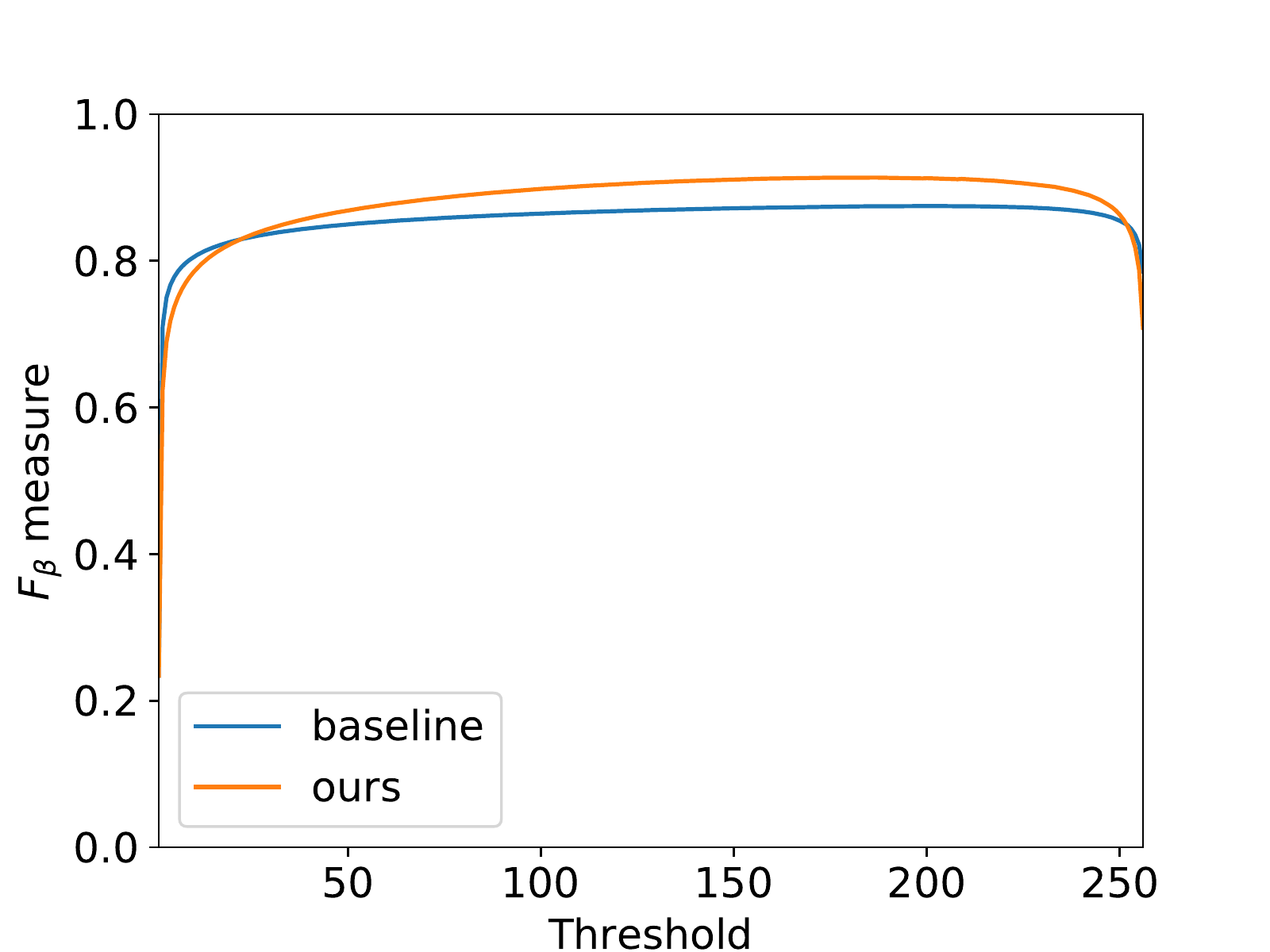}}
\caption{Ablation Study on Threshold-$F_\beta$measure Curves. As shown in the above, the $F_\beta$ measures of our proposed method are considerably higher than those of the baseline, for most thresholds within the range of [50, 200].}
\label{Fig.Fmeasure}
\end{figure}

This section verifies the effectiveness of DNL module and investigates how the number of splits affects the performance of the proposed network. The baseline is built by removing all DNL modules from the proposed network. The baseline has exactly the same architecture as MobileNetV2 but has different input size. We use Precision-Recall curves and Threshold-$F_\beta$ measure curves to compare our proposed method with the baseline. To draw these curves, a list of evenly spaced thresholds is sampled. For each threshold, a tuple of precision, recall and $F_\beta$ measure is calculated, with $\beta^2 = 0.3$. Then we plot the pairs of (recall, precision) in Fig.~\ref{Fig.PRcurve}, and the pairs of (threshold, $F_\beta$ measure) in Fig.~\ref{Fig.Fmeasure}. As shown in Fig.~\ref{Fig.PRcurve} and Fig.~\ref{Fig.Fmeasure}, the proposed module effectively improve the prediction accuracy of the baseline on all three benchmarks.

As TABLE~\ref{table:comp3} displays, Split-9 denotes an accelerated DNL module that divides the input feature tensor into 9 sub-tensors. For Split-$s$ ($s=1, 3, 5, 9$), DNL modules are located after IR-3 and IR-4 shown in Fig.~\ref{fig:DNLnet}. For IR6-split$s$ ($s=1, 5, 10$), a DNL module is inserted after IR-6. As shown in TABLE~\ref{table:comp3}, Split-9 surpasses the baseline by 3.9\% maxF, 1.2\% MAE and 4.1\% S-m on the HKU-IS dataset. The performance of Split-5 is quite close to that of Split-9. Split-9 marginally outperforms Split-1 by 0.11\% maxF and 0.22\% S-measure. For IR6-split$s$, similarly, IR6-split10 exceeds IR6-split1 by 0.9\% maxF, 0.3\% MAE and 0.7\% S-m. The above results suggest that DNL modules effectively improve the baseline and the splits number does not affect much of the prediction quality. Larger splits number could lead to slight improvement, since it helps to maintain better spatial coherence as discussed in Section~\ref{sec:divide-conquer}. If the splits number is larger, then the spatial size of each sub-region becomes smaller. Feature vectors within the same sub-region are more likely to belong to background or the same object. In such cases, non-local pairwise correlations help propagate saliency score of a feature to its adjacent features, which models spatial coherence. Compared with IR6-split1 whose DNL module is at high level, Split-1, whose DNL modules are at middle level, achieves better maxF, MAE and S-m. It suggests that placing the proposed module at middle level better improves the performance.

\begin{table}[!ht]
\caption{The effectiveness of DNL module.}
\centering
\huge
\resizebox{\linewidth}{!}{
    \begin{tabular}{lccccc}
    \toprule
    \multirow{2}{*}{} & \multicolumn{3}{c}{HKU-IS}  & CPUTime                  & \multirow{2}{*}{MAdds} \\
                      & maxF    & MAE     & S-m     & /secs                    &                        \\
    \midrule
    Baseline w/o DNL  & 0.8745  & 0.0572  & 0.8563  & \textbf{0.3906}        & \textbf{9.325B}                  \\
    \midrule
    Split-9           & \textbf{0.9133}  & \textbf{0.0451}  & \textbf{0.8974}  & 0.3993                  & 9.567B         \\
    Split-5           & 0.9133  & 0.0451  & 0.8969  & 0.4019                  & 9.657B                  \\
    Split-3           & 0.9131  & 0.0452  & 0.8967  & 0.4103                  & 9.781B                  \\
    Split-1           & 0.9121  & 0.0458  & 0.8952  & 0.4549                  & 10.42B                  \\
    \midrule
    IR6-split10       & 0.9125  & 0.0459  & 0.8922  & 0.5138                  & 12.25B                  \\
    IR6-split5        & 0.9069  & 0.0479  & 0.8873  & 0.5447                  & 12.90B                  \\
    IR6--split1       & 0.9036  & 0.0489  & 0.8851  & 1.1384                  & 18.13B                   \\
    \bottomrule
    \end{tabular}
}
\label{table:comp3}
\end{table}
The baseline obtains the fastest CPU inference and the lowest MAdds. As the splits number decreases, both CPU time cost and MAdds of the corresponding DNL network become larger. Because smaller splits number results in computing more pairwise similarities. Since CPU time is measured for a whole network, we need to compute \textit{difference} between some network and the baseline to obtain the time cost of DNL modules. For examples, DNL modules in Split-9 additionally costs $399.3-390.6\sim 9$ ms and $9.567-9.325\sim 0.24$B MAdds while DNL modules in Split-1 additionally takes $454.9-390.6\sim 64$ ms and $10.42-9.325\sim 1.1$B MAdds when compared with the baseline model. Compared with Split-1, DNL modules in Split-9 speed up $7\times$ CPU time from 64 ms to 9 ms, and reduce $5\times$ MAdds from 1.1B to 0.24B. For IR6-split$s$, the difference is larger. The DNL module in IR6-split10 reduces $1.1384-0.5138\sim 0.6$ s CPU time and $18.13-12.25\sim 6$B MAdds, in comparison to IR6-split1. The above results indicate that the divide-and-conquer variant of DNL module indeed accelerates the inference. To understand the difference between positioning DNL modules at middle level and high level, we compare Split-1 with IR6-split1. DNL modules are located at the middle level of Split-1 and the high level of IR6-split1. IR6-split1 spends more CPU time (0.7 s) and larger MAdds (7.7B) than Split-1. Thus positioning DNL modules at high level causes more computational cost. Because the time complexity of DNL modules is in proportion to squared number of input channels.
\section{Conclusions}
This paper introduces a novel DNL module that effectively enhances the accuracy of an inverted residual block. A divide-and-conquer variant of DNL is proposed to further accelerate inference. Moreover, we develop a light-weight DNL based network architecture with low memory cost, high inference speed and competitive accuracy. Numeric results declare that our method achieves not only competitive accuracy but also state-of-the-art efficiency among deep CNN based methods.

\ifCLASSOPTIONcaptionsoff
  \newpage
\fi



%



 \bibliographystyle{IEEEtran}
 \bibliography{paper}

\end{document}